%% file: main.tex
\documentclass{article}

\makeatletter
\def\@seccntformat#1{\@ifundefined{#1@cntformat}%
   {\csname the#1\endcsname\quad}  
   {\csname #1@cntformat\endcsname}
}
\let\oldappendix\appendix 
\renewcommand\appendix{%
    \oldappendix
    \newcommand{\section@cntformat}{\appendixname~\thesection\quad}
}
\makeatother
\PassOptionsToPackage{sort,compress,numbers}{natbib}
\usepackage[preprint]{neurips_2023}
\usepackage[utf8]{inputenc} 
\usepackage[T1]{fontenc}    
\usepackage{hyperref}       
\usepackage{url}            
\usepackage{booktabs}       
\usepackage{amsfonts}       
\usepackage{nicefrac}       
\usepackage{microtype}      
\usepackage{xcolor}         

\usepackage{graphicx}
\usepackage{amssymb}
\usepackage{amsmath}
\usepackage{bm}
\usepackage{algorithm}
\usepackage{algpseudocode}
\usepackage{dsfont}
\usepackage{subcaption}
\usepackage[export]{adjustbox}
\usepackage{wrapfig}
\usepackage{multirow}
\usepackage{soul}
\usepackage{authblk}
\usepackage{threeparttable}
\usepackage[normalem]{ulem}
\usepackage[bottom]{footmisc}

\def\*#1{\bm{#1}}
\def\std#1{\footnotesize $\pm$ #1}

\newcommand{\textitbf}[1]{\textbf{\textit{#1}}}

\usepackage{calrsfs}

\newcommand{\params}{R}
\newcommand{\protosample}{R}

\title{Uncertainty-Aware Explanations Through Probabilistic Self-Explainable Neural Networks}

\author[1]{\textbf{Jon Vadillo}}
\author[1]{\textbf{Roberto Santana}}
\author[1,2]{\textbf{Jose A. Lozano}}
\author[3]{\textbf{Marta Kwiatkowska}}
\affil[1]{Department of Computer Science and Artificial Intelligence, \protect\\ University of the Basque Country UPV/EHU}
\affil[2]{Basque Center for Applied Mathematics (BCAM)}
\affil[3]{Department of Computer Science, University of Oxford}
\affil[ ]{\fontfamily{qcr}\selectfont \small
 \{jon.vadillo, roberto.santana, ja.lozano\}@ehu.eus, \protect\\ marta.kwiatkowska@cs.ox.ac.uk
}

\begin{document}

\maketitle

\begin{abstract}
The lack of transparency of Deep Neural Networks continues to be a limitation that severely undermines their reliability and usage in high-stakes applications. Promising approaches to overcome such limitations are Prototype-Based Self-Explainable Neural Networks (PSENNs), whose predictions rely on the similarity between the input at hand and a set of prototypical representations of the output classes, offering therefore a deep, yet transparent-by-design, architecture. In this paper, we introduce a probabilistic reformulation of PSENNs, called Prob-PSENN, which replaces point estimates for the prototypes with probability distributions over their values. This provides not only a more flexible framework for an end-to-end learning of prototypes, but can also capture the explanatory uncertainty of the model, which is a missing feature in previous approaches. In addition, since the prototypes determine both the explanation and the prediction, Prob-PSENNs allow us to detect when the model is making uninformed or uncertain predictions, and to obtain valid explanations for them. Our experiments demonstrate that Prob-PSENNs provide more meaningful and robust explanations than their non-probabilistic counterparts, while remaining competitive in terms of predictive performance, thus enhancing the explainability and reliability of the models.
\end{abstract}

\section{Introduction}
\label{sec:intro}

The complexity in the architectures of state-of-the-art Deep Neural Networks (DNNs) largely accounts for their ``black box'' nature \cite{rudin2019stop}, which is in conflict with one of the basic requirements for trustworthiness \cite{ashoori2019ai}: to be able to explain and understand the decisions of the model. Although several strategies have been proposed in order to explain black box models \cite{zhang2021survey}, they often provide only partial information about their inner workings, such as which parts of the input at hand most condition the output \cite{simonyan2014deep,selvaraju2017gradcam,sundararajan2017axiomatic,bach2015pixelwise,shrikumar2017learning,springenberg2015striving}. However, these approaches do not provide information about how the model processes that information or why they imply the output \cite{rudin2019stop,hase2019interpretable,adebayo2018sanity,lipton2018mythos}. In order to achieve a more transparent classification process, recent works advocate the use of Prototype-Based Self-Explainable Neural Networks (PSENNs, for simplicity) \citep{alvarez-melis2018robust,chen2019this,hase2019interpretable,li2018deep,gautam2022protovae}, which are trained to jointly maximize their prediction performance and their explainability. These approaches rely primarily on prototype-based models, in which the prototypes are learned during the training phase, aiming to capture discriminative and also semantically-meaningful features. In this way, the output classification is based on the similarity between the input and the learned prototypes, providing an intuitive and human-understandable classification process.

On the other hand, current DNNs, including the PSENNs discussed above, are generally optimized with the aim of finding the point estimate values of the parameters that minimize a given loss term, leading to models with a deterministic inference process. An important drawback of these approaches is that, generally, it is not possible to measure to what extent the model is confident in its own prediction, known in the literature as the epistemic uncertainty of the model 
\cite{hullermeier2021aleatoric,kwon2022uncertainty,gal2016uncertainty}. Indeed, it has been shown that DNNs tend to be overconfident in their predictions \cite{guo2017calibration}, classifying inputs with high probability even when random or out-of-distribution inputs are provided. Being unable to detect when the model is making such unreliable or uninformed decisions dramatically reduces the trustworthiness and reliability of the model, and is, therefore, of great concern in high-stakes applications, as recognized in the EU AI Act \cite{aiact}. 
To address this issue, recent works proposed replacing point estimates for the parameters of the model with a probability distribution over their values \cite{goan2020bayesian}. This leads to a stochastic model, for which it is possible to measure the epistemic uncertainty, and, thus, identify when the model is making uncertain predictions that should not be trusted.

In this paper, we aim to address the aforementioned drawbacks and challenges by leveraging recent tools from uncertainty estimation in DNNs, and combining them with PSENN architectures in order to develop \textit{Probabilistic PSENNs (Prob-PSENNs)}. 
The key feature of this model is that the prototypes are defined as random variables, for which suitable probability distributions are learned during the training phase of the model -- jointly with the rest of the parameters of the network. This probabilistic paradigm allows us not only to improve the explainability of the classification process, but also to capture different sources of uncertainty regarding both the output and the explanation. Furthermore, since the output of the PSENNs directly depends on the explanatory component \cite{li2018deep,chen2019this,gautam2022protovae}, we can establish connections between their corresponding uncertainties, enabling a more thorough analysis of the prediction process, and allowing the models to self-explain their own uncertainties.

The main contributions of this work are summarized below:
\begin{enumerate}
    \item We introduce Prob-PSENN, a probabilistic reformulation of PSENNs that enhances trustworthiness by addressing two critical aspects: (i) interpretability and transparency, and (ii) uncertainty quantification.

    \item 
    We demonstrate how the probabilistic reinterpretation of the prototypes substantially enhances the explanatory capabilities of the model, enabling more diverse, meaningful and robust explanations than those achievable with the non-probabilistic counterparts.
    
    \item Prob-PSENNs allow us, for the first time, to model the uncertainty in both the explanations and predictions, which is a missing feature in conventional PSENNs.
    We formalize different explanatory uncertainty notions, and show how Prob-PSENNs can (i) quantify those uncertainties, (ii) explain them, and (iii) leverage them to improve their predictive performance.
    
    \item The effectiveness of Prob-PSENNs is evaluated on several image and tabular data classification tasks, demonstrating their reliability and trustworthiness. 
\end{enumerate}

\section{Preliminaries: Prototype-Based Self-Explainable Neural Networks (PSENNs)}
\label{sec:senn_intro}

Let us consider a classification problem in which the goal is to classify $d$-dimensional inputs $x \in \mathbb{R}^d$ into one of the $c$ possible classes in $Y=\{y_1,y_2,\dots,y_c\}$. Building on the frameworks introduced in \cite{li2018deep, gautam2022protovae}, the core components of the PSENNs are described below. We refer the reader to Appendix~\ref{app:related_work} for a justification of the choice of these frameworks, as well as a discussion of related works.

\begin{figure}[]
    \centering
    \includegraphics[scale=0.48]{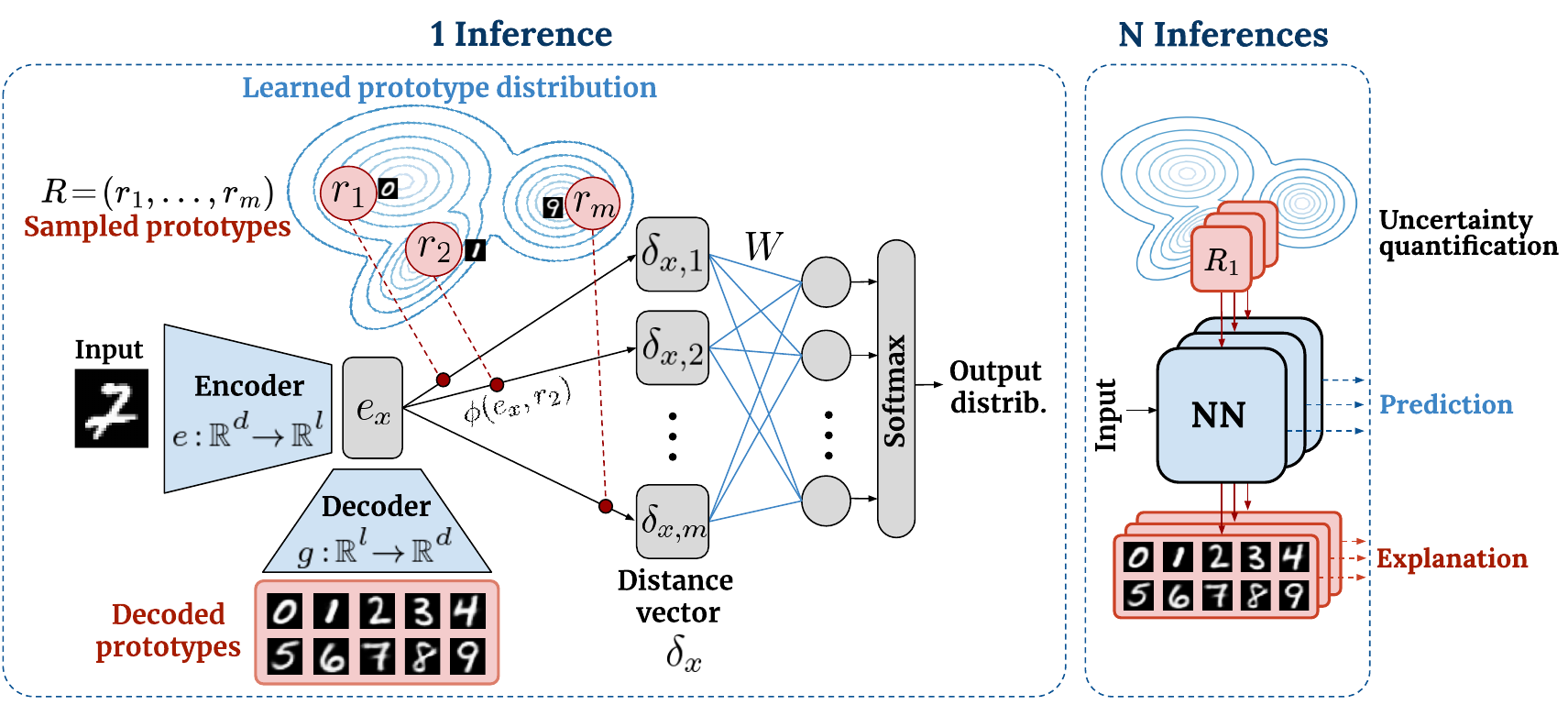}
    \caption{Architecture of the proposed Probabilistic PSENN (Prob-PSENN). The model replaces deterministic prototypes with probabilistic prototype distributions, enabling the sampling of diverse yet representative prototypes from the latent space. To account for prototype variability, $N$ samples of the prototypes can be considered during the inference process, producing diverse predictions that enable capturing both explanatory and predictive uncertainties.    
    }
    \label{fig:BSENN_architecture}
\end{figure}

The first component of these architectures is an encoder module $e : \mathbb{R}^d \rightarrow \mathbb{R}^l$, where $l$ denotes the dimensionality of the \textit{latent} space to which the input is encoded. 
The latent representation computed by the encoder $e(\cdot)$ for the input $x$ will be denoted as $e_x = e(x)$. Conversely, let $g : \mathbb{R}^l \rightarrow \mathbb{R}^d$ denote a decoding module, capable of reconstructing the original input $x$ from the corresponding encoded representation $e_x$.
The key component of PSENNs involves the generation of $m$ prototypes in the latent space: $R=(r_1, r_2, \dots, r_m)$, with $r_i \in \mathbb{R}^l$, $1\leq i \leq m$. 
The aim of generating these prototypes is to capture diverse, discriminative and semantically meaningful features capable of representing the classes of the problem.
In this way, the similarity between an encoded input $e_x$ and each of the $m$ prototypes can be employed to determine the output class for $x$ in an easily interpretable manner, based on the assumption that an input $x$ belonging to class $y_i$ will be closer to those prototypes representing the class $y_i$ than to the prototypes corresponding to the remaining classes.
To formalize the aforementioned similarities, let the column-vector $\delta_x=(\delta_{x,1}, \delta_{x,2}, \dots, \delta_{x,m})^\intercal$ represent the distance between an encoded input $e_x$ and the $m$ prototypes in $R$. More specifically, the distance between $e_x$ and the $i$-th prototype $r_i$ will be computed as
$
    \delta_{x,i} = \phi(e_x, r_i), \ 1 \leq i \leq m,
$
based on a distance metric $\phi : \mathbb{R}^l \times \mathbb{R}^l \rightarrow \mathbb{R}$. The distance vector $\delta_x$ will condition both the output classification and the corresponding explanation, as detailed below.

\paragraph{Classification}
\label{sec:senn_intro_class}
Taking $\delta_x$ as the input, the output provided by the model will be determined by a prototype classifier $h : \mathbb{R}^m \rightarrow [0,1]^c$. As in \cite{li2018deep}, a classification network
$h(\delta_x)=s(W\delta_x)$
will be assumed, where $W$ is a $c \times m$ weight matrix and $s(\cdot)$ the \textit{softmax function}.
For simplicity, the end-to-end classification process will be denoted as $f(x) = h(\delta_x)$. Notice that $f(x)$ outputs a vector representing the probability with which the input belongs to each of the possible classes of the problem, according to the model. The probability assigned to each class $y_i$ will be denoted as $f_i(x)$, $1\leq i \leq c$, and thus we can define the output as:
$    f(x) = \bigl(f_1(x), f_2(x), \dots, f_c(x)\bigr)$.

\paragraph{Explanation}
\label{sec:senn_intro_expl}
Since the output class is based on the distance-vector $\delta_x$, the explanation $\xi_x$ for the prediction $f(x)$ can be determined also by the similarity between the prototypes and the input at hand:
$
    \xi_x = 
    \bigl\{ 
        \{g(r_1), \delta_{x,1}\}, 
        \{g(r_2), \delta_{x,2}\},
        \dots, 
        \{g(r_m), \delta_{x,m}\}
    \bigr\},
$
where $g(\cdot)$ is the decoder defined above.
The interpretability of the model relies, therefore, on to what extent the decoded prototypes are capable of capturing relevant and semantically-meaningful features representing each class of the problem, which is enforced during the training phase by means of including interpretability regularizers in the loss function \cite{li2018deep,gautam2022protovae}.

\section{Probabilistic Prototype-Based Self-Explainable Neural Networks (Prob-PSENNs)}
As mentioned above, PSENNs treat the prototypes as network parameters, for which point estimate values are computed and hence are fixed once the training is completed.
In this section, we introduce Prob-PSENNs, which generalize the prototypes by means of defining and learning a probability distribution over their values, as we illustrate in Figure \ref{fig:BSENN_architecture}.
The main motivation for this probabilistic formulation is discussed below.

First, for each class of the problem, there might exist several sets of prototypes $R$ capable of representing the data with comparable effectiveness. For illustration, let us consider a handwritten digit classification scenario, and the fact that there are several ways of representing each digit. This implies that there are several representative prototypes for each class. Consequently, instead of relying on a single set of prototypes, it might be more reasonable to consider all those possible variations in order to classify an input. 
Furthermore, in previous approaches, the number of prototypes $m$ needed to be finite and defined beforehand, even if, in practice, it might be cumbersome or infeasible to determine how many prototypes will be required to capture sufficiently diverse characterizations of the classes.
In contrast, our probabilistic reformulation enables a more flexible way to decide the variety of the prototypes required to represent the classes, which will be determined, assuming one prototype distribution per class, by the variance of the distributions. 

Finally, the choice of the prototypes in $R$ conditions not only the output but also the explanation. Therefore, placing a probability distribution over the prototypes allows us to enable a novel probabilistic framework for the generation of the explanation as well, which can be employed to capture different sources of uncertainty in the explanations of the model, and thereby produce more informative and useful explanations.

\subsection{Predictive distribution}
Instead of considering a point estimate for $\params$, we will use a random variable, denoted $\*\params$. 
In this way, we can consider $m=c$ random class-prototypes, such that $\*r_i$ represents the class $y_i$, $1\leq i \leq c$. This eliminates the need to estimate $m$ and avoids having to restrict the model to a fixed number of prototypes per class, as each $\*r_i$ will model the required variability within the corresponding class. Due to the randomness in the network's prototypes, the inference of the model will be stochastic, and will be denoted as $f^{\*\params}(x)$. 
In this way, a predictive distribution for the output can be defined:
\begin{equation}
\label{eq:predictive_distr}
    p(\*y | x) = 
    \mathbb{E}_{ p(\*\params) }[p(\*y|x,\*\params)] =
    \int p(\*y| x, \*\params) \cdot p(\*\params) \ d\*\params = \int f^{\*\params}(x) \cdot p(\*\params) \ d\*\params.
\end{equation}
In practice, $p(\*y | x)$ can be approximated by considering a finite number of $N$ samples from $p(\*\params)$:
\begin{equation}
\label{eq:pred_distr_approx}
\tilde{p}(\*y | x) = \bar{f}(x)=\frac{1}{N}\sum_{n=1}^N f^{\protosample_n}(x),  \ \ \text{with} \ \ \protosample_n \sim p(\*\params), \ 1\leq n \leq N.
\end{equation}

\subsection{Capturing the predictive uncertainty}
\begin{figure}[!t]
    \centering
    \includegraphics[scale=0.35]{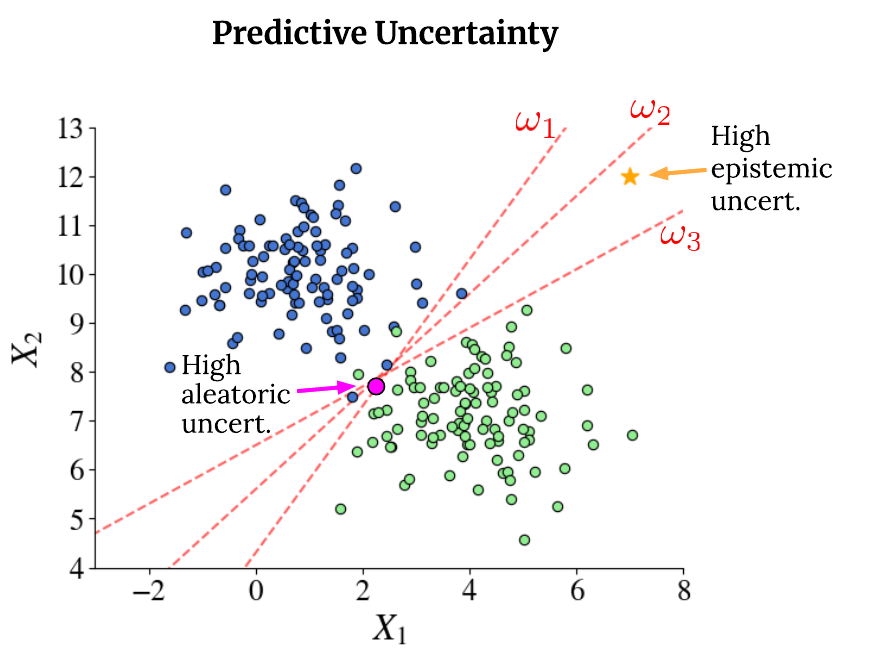}
    \hspace{1.0cm}
    \includegraphics[scale=0.35, trim=150 92 0 0, clip]{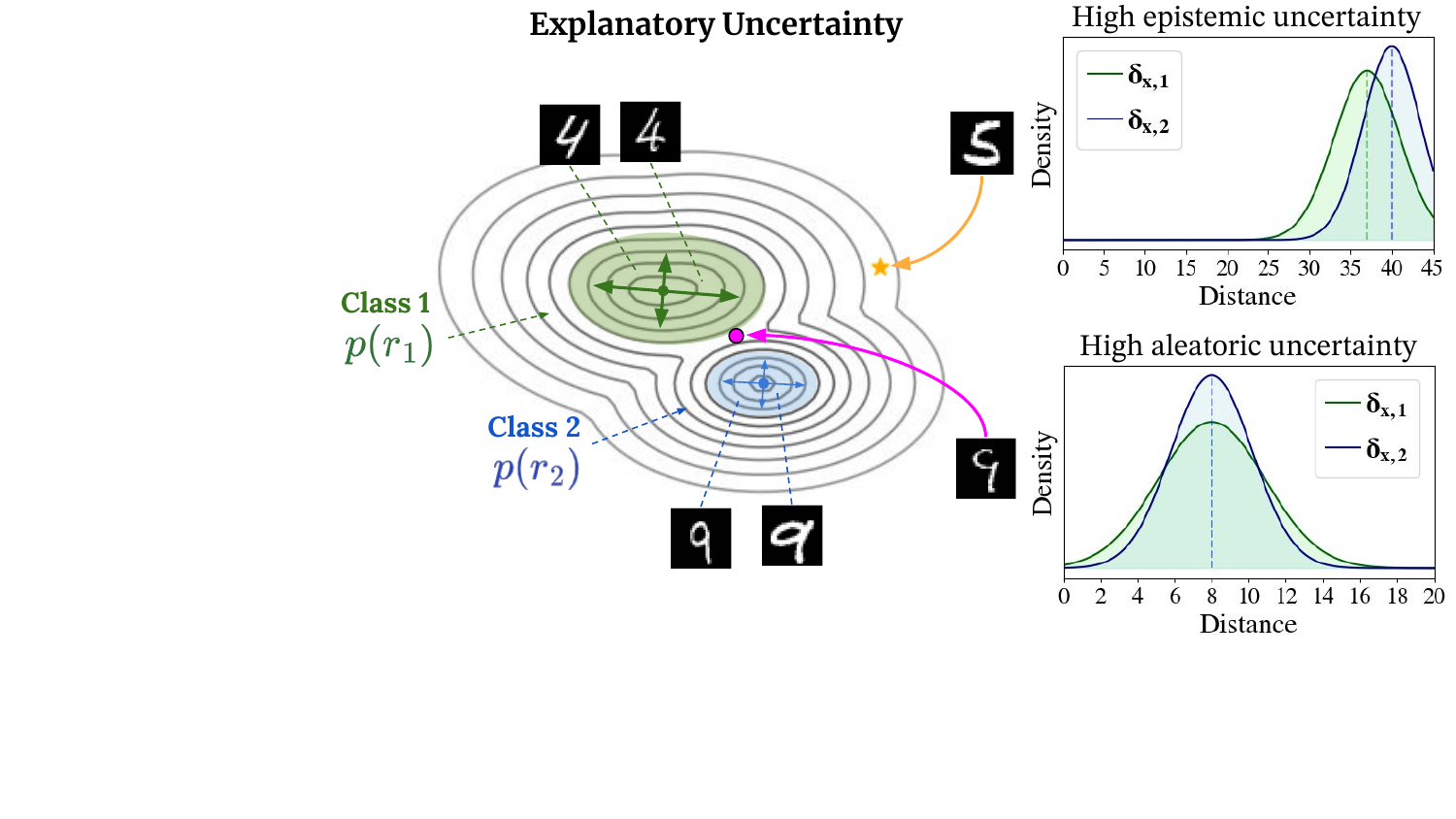}
    \caption{
    Comparison between the predictive and explanatory uncertainty, illustrated for a binary classification problem with a two-dimensional input space. In the left subfigure, dashed lines represent different decision boundaries. In the right subfigure, colored regions represent high density regions for the prototype distribution $p(r_1)$ and $p(r_2)$. 
    }
    \label{fig:output_uncert_illust}
\end{figure}

The predictive distribution defined in Equation \eqref{eq:predictive_distr} allows us to better model the \textit{predictive uncertainty} of a Prob-PSENN compared to its deterministic counterpart.
First, the total predictive uncertainty can be measured by means of the entropy \cite{shannon1948mathematical} of the prediction:
${H[p(\*y|x)] =
-\sum_{i=1}^c p(\*y=y_i|x) \cdot \log p(\*y=y_i|x)}$.
However, for a model with fixed (pointwise) values for the parameters, the source of this uncertainty will be only \textit{aleatoric} uncertainty, which refers to the possibly unavoidable noise, randomness or ambiguity inherent in the data.
Considering a probability distribution for the parameter values instead of point estimates allows us to also capture the 
epistemic uncertainty \cite{hullermeier2021aleatoric, kendall2017what, kwon2022uncertainty}, which accounts for the model uncertainty about the true parameters modeling the data (e.g., as a consequence of insufficient training data). This enables a safer use of the model in critical tasks, since it allows us to identify when the model is making unreliable predictions.
An illustrative comparison between aleatoric and epistemic uncertainty is provided in Figure~\ref{fig:output_uncert_illust} (left).

Thus, for the predictive distribution defined in Equation~\eqref{eq:predictive_distr}, $H[p(\*y|x)]$ encapsulates both aleatoric and epistemic uncertainty. To measure epistemic uncertainty only, the mutual information between the output $\*y$ and the parameters $\*\params$ can be employed \cite{houlsby2011bayesian,gal2016uncertainty,hullermeier2021aleatoric,depeweg2018decomposition}, which, in our case, can be formalized as:
$
I(\*y, \*\params | x)
= 
H[\,p(\*y|x)] - \mathbb{E}_{p(\*\params)}\big[H[\,p(\*y|x,\*\params)]\big], 
$
while $\mathbb{E}_{p(\*\params)}\big[H[\,p(\*y|x,\*\params)]\big]$ 
represents a measure of aleatoric uncertainty. Notice that the values of all these metrics can be normalized to the range $[0,1]$, where $1$ represents the maximum uncertainty, by dividing the resulting value by the entropy of the uniform categorical distribution of $c$ categories. In practice, $H[p(\*y|x)]$ can be approximated by $H[\tilde{p}(\*y|x)]$ (see Equation~\ref{eq:pred_distr_approx}).

\subsection{Capturing the explanation uncertainty}
\label{sec:explanation_uncertainty}

\begin{figure}
    \centering
    \includegraphics[scale=0.42, trim=0 95 0 0, clip]{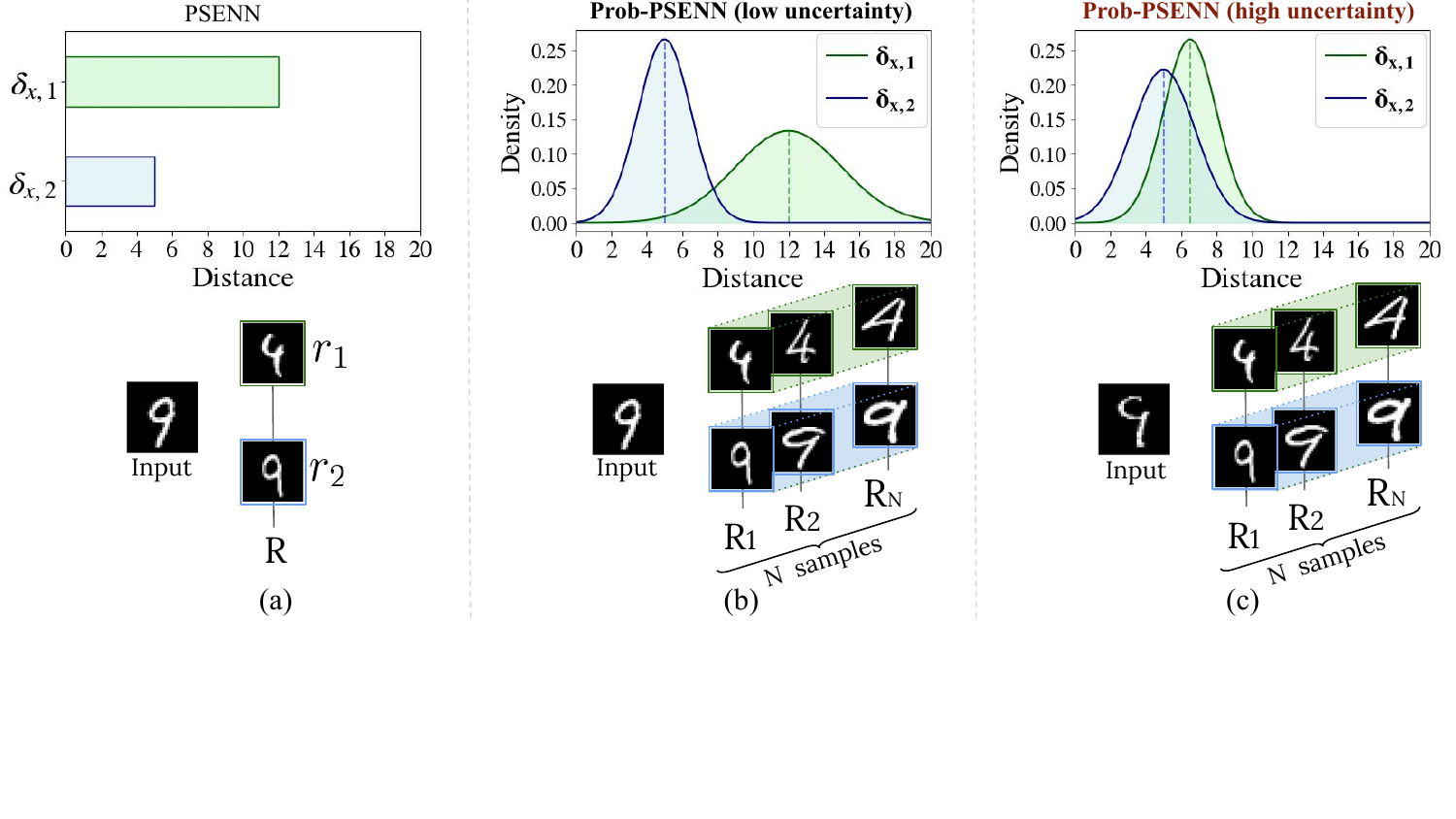}
    \caption{
    Illustrative comparison between a conventional PSENN (left column) and the proposed Probabilistic-PSENN (middle and right columns) regarding the computation of the distance-vector $\delta_x$. Whereas a fixed set of prototypes $R$ is used in (a), leading to a point estimate $\delta_x$, placing a probability distribution over $\*R$, as in (b) and (c), induces a probability distribution over the distance $\delta_{x,i}$ between the encoded input and the random prototypes $\*r_i$ representing the class $y_i$, $1 \leq i \leq c$.
    }
    \label{fig:delta_distrib}
\end{figure}

With a PSENN with fixed prototypes, only a point estimate $\delta_x$ can be provided for the input at hand, that is, a point estimate on how close the input is to the particular prototypes selected to represent each class, as illustrated in Figure~\ref{fig:delta_distrib}-(a).
In contrast, the probability distribution over the prototypes considered in Prob-PSENNs induces a probability distribution over each value of $\delta_x$,
$\, p(\*\delta_{x,i}) \! = \! p\bigl( \phi(e_x, \*r_i) \bigr)$,
as exemplified in Figure~\ref{fig:delta_distrib}-(b) and Figure~\ref{fig:delta_distrib}-(c).
This allows us to compute not only more representative information for each value of $\delta_x$, but also to compute the uncertainty in the explanation, as will be detailed below. Furthermore, this enables a direct and explicit connection between the explanation uncertainty and the output uncertainty, which is a key advantage of Prob-PSENNs.

\subsubsection{Types of explanation uncertainty}
Similarly to the predictive uncertainty, different types of uncertainty can be attributed to the explanations of Prob-PSENNs.
In the context of explanations, we will refer to \textit{aleatoric} uncertainty as the uncertainty in determining which class-prototypes best represent the input at hand, that is, in determining which prototypes are closest to the encoded input, possibly due to ambiguity in the representation of the input. Therefore, high aleatoric uncertainty will imply highly overlapping distance-distributions $p(\*{\delta}_{x,i})$.
In contrast, \textit{epistemic} explanatory uncertainty will be high for those inputs that lie far from the training distribution, and, consequently, also far from the distribution over the prototypes, resulting in uninformative and non-representative explanations for that input. Therefore, high epistemic uncertainty will imply large distances with respect to all the prototypes.  
An illustrative representation of these types of explanatory uncertainty is provided in Figure~\ref{fig:output_uncert_illust} (right).

\subsubsection{Quantifying explanation uncertainty}
Given that, in the general case, computing $p(\*\delta_{x,i})$ will be intractable both analytically and numerically, we will rely on sample-based estimators in order to compute meaningful information about those distributions.  
To begin with, assuming $\delta_{x}^{(1)}, \dots, \delta_{x}^{(N)}$ are the distance-vectors obtained for $N$ inferences of the model (i.e., for $N$ random sets of prototypes), 
the sample mean 
${
\bar{\delta}_{x,i} 
= \frac{1}{N}\sum_{n=1}^N \delta_{x,i}^{(n)}}$
can inform us about the average similarity between the input and the prototypes of the class $y_i$, while the sample variance \text{Var} 
${\bigl( p(\*\delta_{x,i}) \bigr) \approx \frac{1}{N-1} \sum_{n=1}^N 
\bigl( \delta_{x,i}^{(n)} - \bar{\delta}_{x,i} \bigr)^2}$
can be considered a measure of uncertainty about those similarities. 
    
While the aforementioned two metrics describe each distribution $p(\*\delta_{x,i})$ separately, considering all these distributions jointly allows us to estimate aleatoric and epistemic uncertainty in the explanations.
As mentioned earlier, aleatoric uncertainty can be determined based on the overlap between the probability density functions describing $p(\*\delta_{x,i})$, which will be denoted as $f_{\*\delta_{x,i}}(\cdot)$. For a binary case, the overlap can be measured by means of (normalized) overlapping indices 
\cite{pastore2019measuring}:
$
    		\int_{\mathbb{R}} 
        \min [ 
        f_{\*\delta_{x,1}}\!(z), \, f_{\*\delta_{x,2}}(z) 
        ] \, dz. \ 
$
In order to address multi-class problems, we will generalize the metric as the maximum overlap between the density function $f_{\*\delta_{x,i^*}}(\cdot)$ corresponding to the most likely class $y_{i^*}$ and the functions corresponding to the remaining classes:
    \begin{equation}
    \label{eq:expl_aleatoric}
    \hspace{1cm}
    \mathcal{U}_A(x) =
        \max_{
        		\substack{
        			j=1,\dots,c \\
        			j\neq i^*
        		}
        }
        \left[ 
          \int_{\mathbb{R}} 
          \min [ 
        		f_{\*\delta_{x,i^*}}\!(z), \,
        		f_{\*\delta_{x,j}}(z) 
          ] \, dz
        \right].
    \end{equation}
In practice, the density functions can be estimated by means of kernel-density estimation methods, based on the distances obtained for a finite number of inferences $N$.

Regarding epistemic explanatory uncertainty, as explained above, high uncertainty will be achieved for those inputs for which the distances to all the prototypes are large. On that account, the minimum sample mean: $\bar{\delta}_{x,i*} = \min_{1\leq j\leq c} \bar{\delta}_{x,j}$ represents a suitable base metric to evaluate this uncertainty. Furthermore, in order to determine the severity of the uncertainty in a normalized and scale-independent manner, a quantile-based metric will be defined, close to the approach proposed in \cite{petrov2022robustness}. This metric will be based on comparing the value at hand with the values obtained for the data distribution $\mathcal{D}$, which can be estimated, in practice, using a training set $D$:
\begin{equation}
\label{eq:expl_epistemic}
\mathcal{U}_E(x) = 
P_{\!
	\substack{ (x^\prime\!,y)\sim \mathcal{D} \\ \! y=y_{i*} }
}
\big(
	\bar{\delta}_{x^\prime,i*} \leq \bar{\delta}_{x,i*} 
\big).
\end{equation}

\subsection{Creating more meaningful explanations}

In practice, we expect to sample $N$ different explanations, and, thereby, $N$ different yet representative prototypes for each class $y_i\in Y$ (assuming $m=c$). Although presenting all these prototypes as part of the explanation might be impractical, different strategies can be considered to summarize or distill the results, for the sake of more meaningful and useful explanations. 
First, among the prototypes corresponding to the most likely class, the ones closest to the input at hand can be considered the \textit{best candidates} to justify the prediction of the model.
Furthermore, the closest prototypes corresponding to the remaining classes can be taken as \textit{counterarguments} against the most likely class, which can be used, for instance, to identify and explain why an input can be considered ambiguous or challenging to classify. 
On the other hand, the farther prototypes corresponding to the most likely class can be used to exemplify alternative yet still valid representations for that class, which can provide useful complementary information, and is thus  
 valuable for \textit{knowledge-acquisition} purposes.

\subsection{Training procedure}

Let $p_\lambda(\*R)$ represent a parametric probability distribution over the prototypes. The goal of the training procedure will be to optimize the parameters $\lambda$, jointly with the rest of the parameters of the model, in order to maximize both the predictive performance and interpretability of the model. For simplicity, the prototypes $\*r_1\dots, \*r_c$ will be assumed to be independent random variables, with each $\*r_i$ distributed according to a distribution $p_{\lambda_i}(\*r_i)$, $1 \leq i \leq c$, 
and, hence,
$ 
  p_\lambda(\*R) = p_\lambda(\*r_1, \dots, \*r_c)
  = 
  \prod_{i=1}^c p_{\lambda_i}(\*r_i)
$.
The following generalized loss function will be employed to train the entire architecture:
\begin{equation}
\notag
\mathcal{L} = \tau_1 \cdot \mathcal{L}_{NLL} + \tau_2 \cdot \mathcal{L}_{REC} + \tau_3  \cdot \mathcal{L}_{INT}
\end{equation}
where,
\begin{equation}
\notag
\mathcal{L}_{\text{NLL}}
\! = \! 
-
\frac{1}{N} 
\sum_{n=1}^N 
\sum_{(x,y)\in D}
\sum_{i=1}^c  \mathds{1}(y\!=\!y_i)  \cdot \log\bigl(  f_i^{R_n}(x)  \bigr), 
\ \ \,
R_n \!=\! [r_i]_{i=1}^c \text{ s.t. } r_i \! \sim \! p_{\lambda_i}(\*r_i), \ \, 1\!\leq\! n \!\leq\! N,
\end{equation}
\begin{equation}
\notag
\mathcal{L}_{\text{REC}} = 
\sum_{(x,\cdot)\in D} \!\!
|| x - g\bigl(e(x)\bigr)||_2^2,
\quad \quad \quad \quad \quad \quad
 \mathcal{L}_{\text{INT}} = 
 -  \!
 \sum_{(x,y)\in D} \,
 \sum_{i=1}^c \mathds{1}(y=y_i)  \cdot \log p_{\lambda_i}\bigl(e(x)\bigr),
\end{equation}
and $\tau_1, \tau_2, \tau_3 \in \mathbb{R}^+$ weight the contribution of each term. Notice that $\mathcal{L}_{\text{NLL}}$ represents the classification (cross-entropy) loss, averaged for $N$ samples of prototypes, $\mathcal{L}_{\text{REC}}$ the reconstruction loss, and $\mathcal{L}_{\text{INT}}$ an interpretability loss,
as it will encourage the distribution over the prototypes to be similar to the distribution of the encoded data, which is a crucial requirement for interpretability \cite{li2018deep}. Based on this methodology, the entire architecture can be jointly optimized by means of conventional gradient descent approaches.

\section{Experiments}
\label{sec:results}

In this section, we describe the experimental setup designed to evaluate the effectiveness and capabilities of the proposed model, followed by a presentation and discussion of the corresponding results. The primary goal of these experiments is to demonstrate how Prob-PSENNs can: (i) learn highly diverse and prototypical representations of the classes, (ii) provide reliable uncertainty quantification for both the explanation and the classification, and (iii) leverage the uncertainty quantification mechanism to improve the accuracy, reliability and trustworthiness of the model.

\subsection{Setup}

The effectiveness and capabilities of the proposed model will be assessed mainly on the MNIST \cite{lecun1998gradientbased},  Fashion-MNIST \cite{xiao2017fashionmnist}, Kuzushiji-MNIST (also known as K-MNIST) \cite{clanuwat2018deep} and SVHN \cite{netzer2011reading} datasets. This choice of datasets is based on the fact that they provide a clear and coherent notion of prototypes to facilitate the assessment of our results, particularly regarding explanation quality (e.g., prototype diversity) and uncertainty quantification (e.g., ambiguity), while offering increasing complexity in the representations (from digits to visually complex handwritten Hiragana characters). 
Additional experiments will be carried out for the EMNIST \cite{cohen2017emnist} dataset, in order to assess the scalability of our model to a larger number of classes, and the Sensorless Drive Diagnosis dataset \cite{bayer2013sensorless}, in order to illustrate the applicability of Prob-PSENN to tabular data.

The autoencoder module of Prob-PSENN will be implemented by a CNN-based Variational AutoEncoder (VAE) for image datasets, which adds an additional regularization term to the overall loss function, namely, the Kullback–Leibler (KL) divergence between the learned latent distribution and a prior (Gaussian) distribution \cite{kingma2014autoencoding}. 
For tabular data, an MLP architecture will be employed for the autoencoder. The full implementation details are reported in Appendix~\ref{app:implementation}. For each prototype distribution $p_{\lambda_i}(\*r_i)$, $1\leq i \leq c$, a full covariance Gaussian distribution will be assumed, whose parameters (mean vectors and covariance matrices) are optimized jointly with the rest of the network parameters. Covariance matrices are parameterized via Cholesky decomposition to ensure positive definiteness. As in \cite{li2018deep}, the squared Euclidean distance will be used as the distance metric $\phi$ to measure the distance between the encoded input and the sampled prototypes (in the latent space). The final classification layer $W$ will be set as $W=-I$, where $I$ represents the identity matrix of dimension $c$, thus ensuring that the association between the prototype $\*r_i$ and the class $y_i$, $1\leq i \leq c$, is explicit and unequivocal, further promoting transparency. The Adam optimization method \cite{kingma2015adam} has been used to train the networks, using a learning rate of $0.001$. Our code is publicly available at \url{https://github.com/vadel/Prob-PSENN}.

\subsection{Learned latent spaces and distributions over the prototypes}

\begin{figure}[t]
    \centering
    \begin{subfigure}[b]{0.49\textwidth}
         \centering
         \includegraphics[scale=0.48]{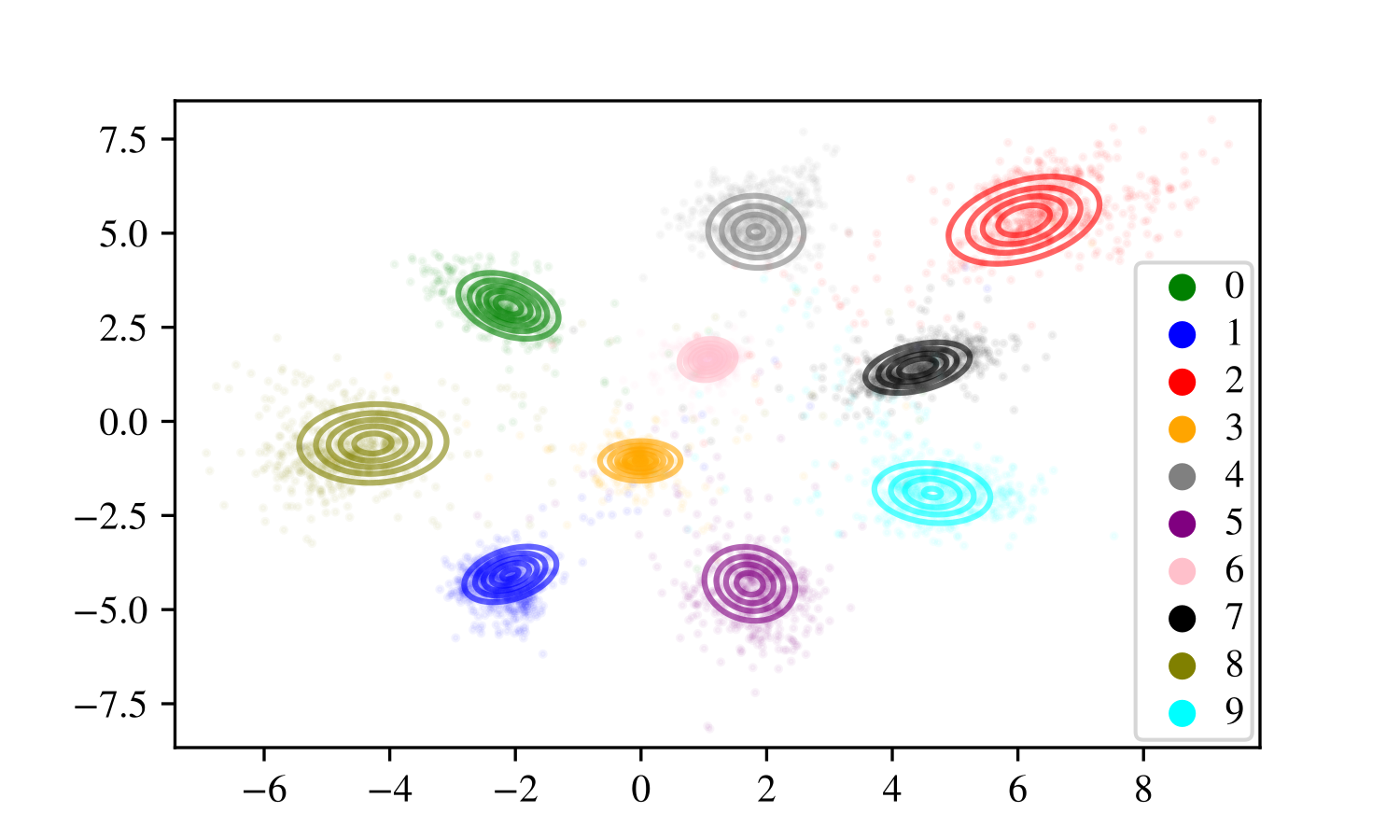}
         \caption{Latent space.}
         \label{fig:mnist_2d_summary_space}
    \end{subfigure}
    \hfill
    \begin{subfigure}[b]{0.49\textwidth}
         \centering
         \includegraphics[scale=0.48]{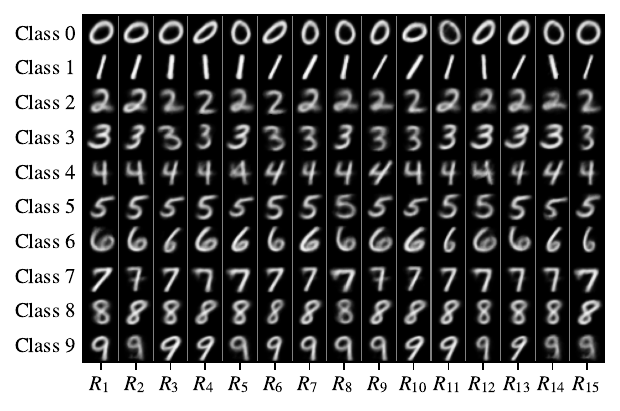}
         \caption{Randomly sampled sets of prototypes.}
         \label{fig:mnist_2d_summary_protos}
    \end{subfigure}
    \caption{
    Illustrative results obtained with a Prob-PSENN with $l=2$ for the MNIST dataset.
    (a)~Learned latent space, where the contours represent the prototype distributions $p_{\lambda_i}(\*r_i)$ and the dots represent encoded input samples.
    (b) 15 random sets of prototypes 
    $R_n = (r_1,\dots,r_c) \text{ s.t. }  r_i \sim p_{\lambda_j}(\*r_j),  1 \!\leq\! i \!\leq\! c$, $1 \! \leq \! n \! \leq \! 15$.
    }
    \label{fig:mnist_2d_summary}
\end{figure}

First, for illustration, the results obtained for the case in which $l=2$ (i.e., a two-dimensional latent space is considered) are shown in Figure~\ref{fig:mnist_2d_summary}, for the MNIST dataset. The left subfigure shows the learned latent space, visualizing the latent representations assigned to a random set of test data (dots), as well as the learned distribution over the prototypes $p_{\lambda_i}(\*r_i)$, $1\leq i \leq c$ (contour lines). 
The right subfigure represents 15 random sets of prototypes $R$ sampled from the learnt distributions, and decoded by the decoder $g$. As can be seen, even for a very constrained (low-dimensional) latent space, the model is capable of capturing and decoding prototypical and varied representations for each class.

\begin{figure}[]
    \centering
    \includegraphics[scale=0.5]{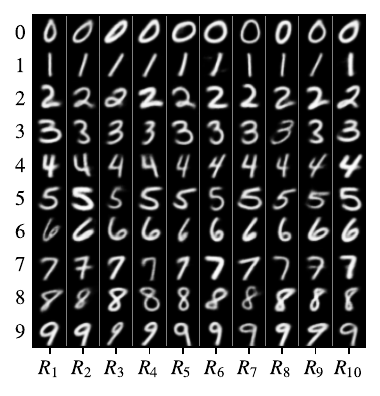}
    \includegraphics[scale=0.5]{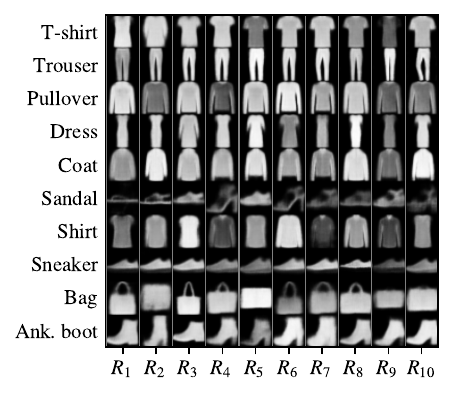}
    \includegraphics[scale=0.5]{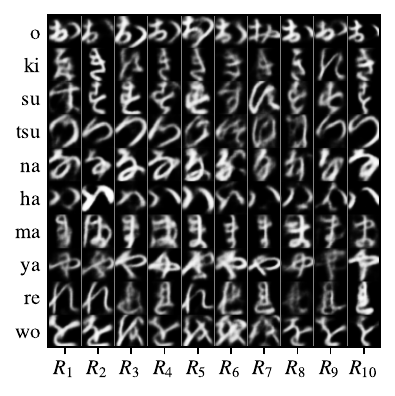}
    \includegraphics[scale=0.5]{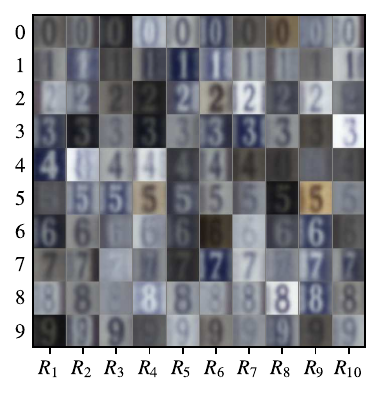}
    \caption{
    Visualization of prototypes sampled from the  distributions learned with Prob-PSENNs ($l=40$), for the MNIST, F-MNIST, K-MNIST and SVHN datasets.
    }
    \label{fig:prototypes_all_datasets}
\end{figure}

Figure \ref{fig:prototypes_all_datasets} shows randomly sampled sets of prototypes for the four datasets considered, for Prob-PSENNs trained with a latent dimensionality of $l=40$. As can be observed, the increase in the dimensionality of the latent space makes it possible to capture prototype distributions with a higher degree of complexity in their structure, and to obtain more detailed decodings. 
Furthermore, even for complex datasets such as K-MNIST (see Appendix \ref{app:kmnist}), the learned prototype distributions are able to capture representative, high-quality and diverse characterizations of the classes. More results are shown in Appendices \ref{app:mnist_results}, \ref{app:fashion} and \ref{app:kmnist}.

\subsection{Evaluating and illustrating the explanatory uncertainties}
\begin{figure}[!h]
    \centering
    \includegraphics[scale=0.33]{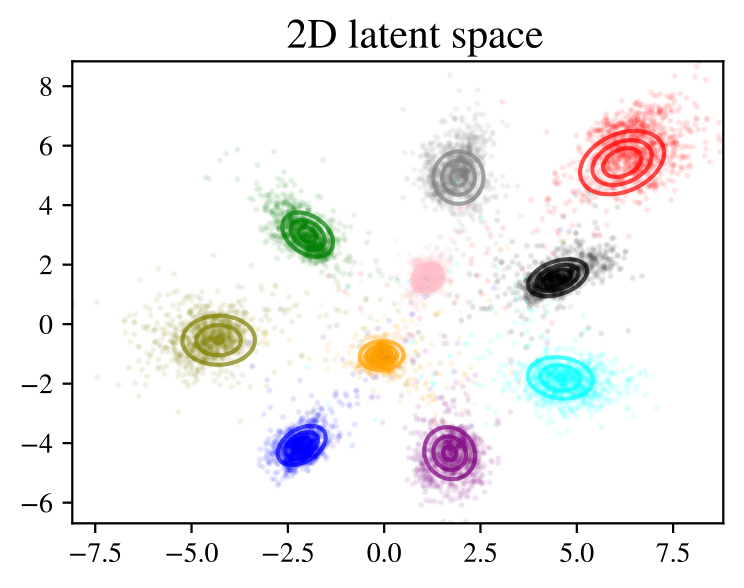}
    \includegraphics[scale=0.33]{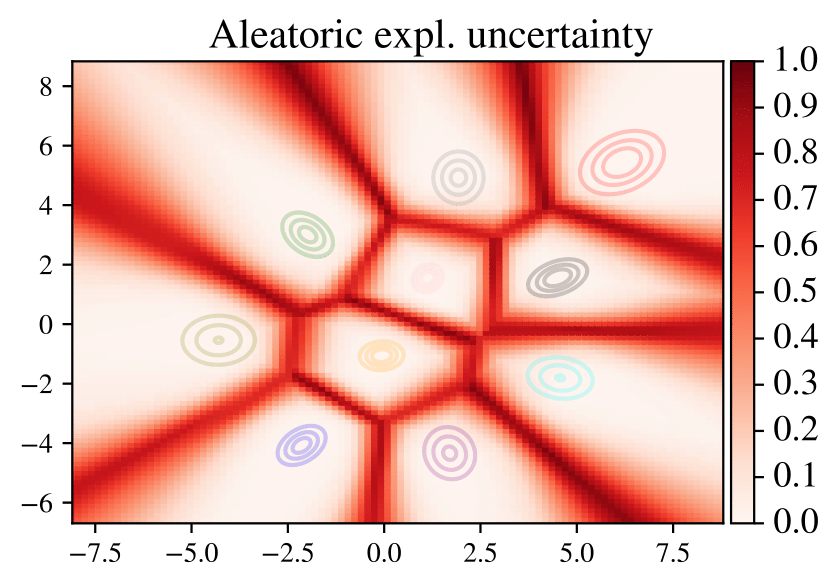}
    \includegraphics[scale=0.33]{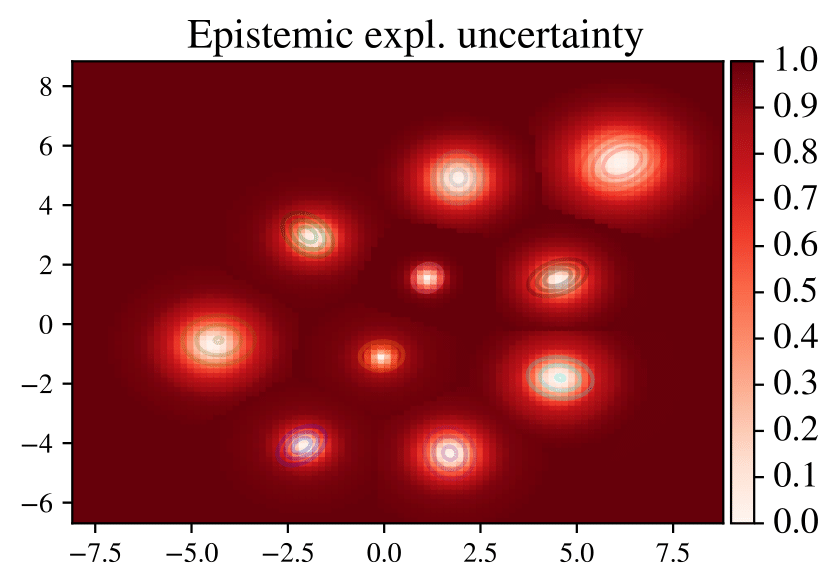}
    \includegraphics[scale=0.33]{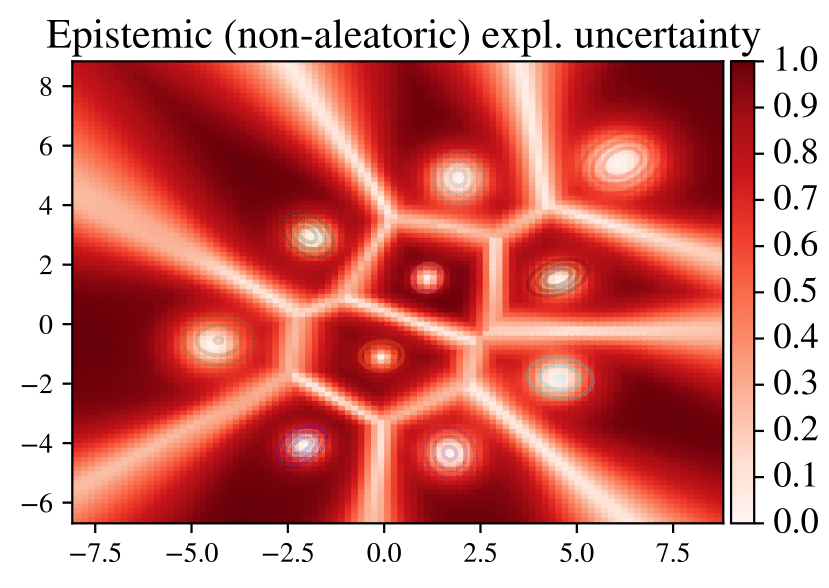}
    \caption{Evaluating the explanatory uncertainty in different regions of the latent space learned by Prob-PSENN.
    }
    \label{fig:uncertainty_regions}
\end{figure}

Figure~\ref{fig:uncertainty_regions} compares, for a Prob-PSENN with $l=2$ and the MNIST dataset,
the areas of the latent space with high explanatory uncertainty. 
The aleatoric and epistemic explanatory uncertainties (second and third columns) have been evaluated using the metrics $\mathcal{U}_A(x)$ and $\mathcal{U}_E(x)$ defined in Equations~\eqref{eq:expl_aleatoric} and \eqref{eq:expl_epistemic}, respectively. 
The fourth column shows the areas solely with high epistemic uncertainty (i.e., with no aleatoric uncertainty), which has been computed as $\mathcal{U}_E(x) - \mathcal{U}_A(x)$. As can be seen from the results, regions with high aleatoric uncertainty correspond to areas in the latent space that lie between distinct class clusters.
On the other hand, the areas with high epistemic uncertainty are those far from the regions to which the prototype distributions assign high density, and, therefore, the explanations are no longer representative of those regions.

\begin{figure}[t]
    \centering
    \includegraphics[scale=0.43]{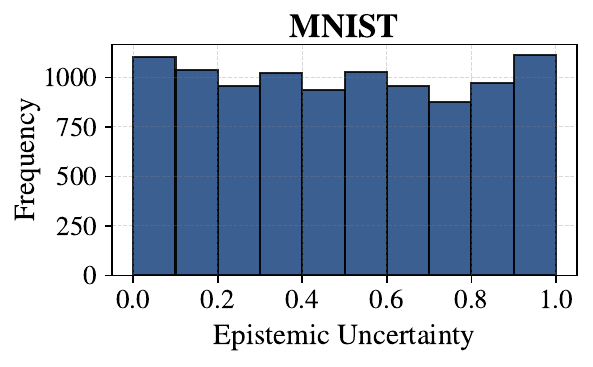}
    \ \ 
    \includegraphics[scale=0.43]{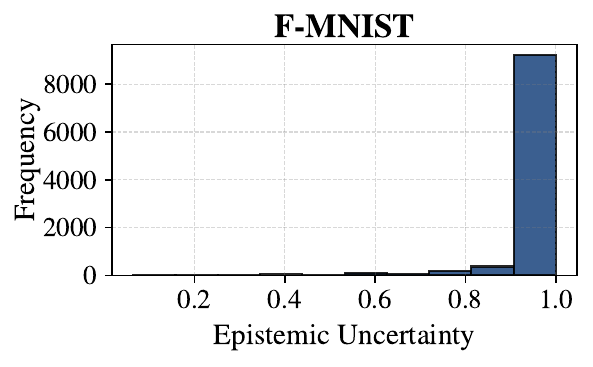}
    \ \ 
    \includegraphics[scale=0.43]{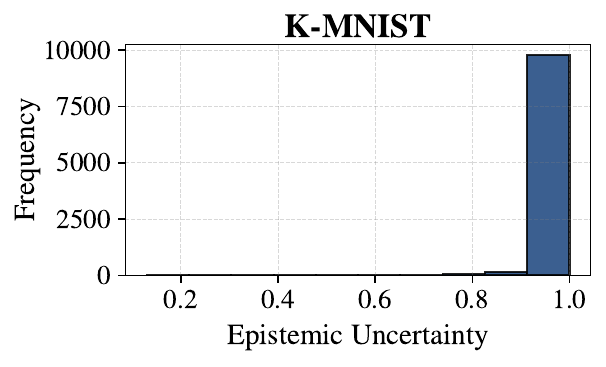}
    \caption{
    Histograms showing the explanatory epistemic uncertainty obtained by a Prob-PSENN (trained on the MNIST dataset) for 10000 test samples from MNIST, F-MNIST and K-MNIST. Notice that the scales of the Y-axis differ in the three plots.
    }
    \label{fig:mnist_high_epistemic}
\end{figure}

\begin{figure}[t]
    \centering
    \begin{subfigure}{\textwidth}
    \centering
    \includegraphics[scale=0.28]{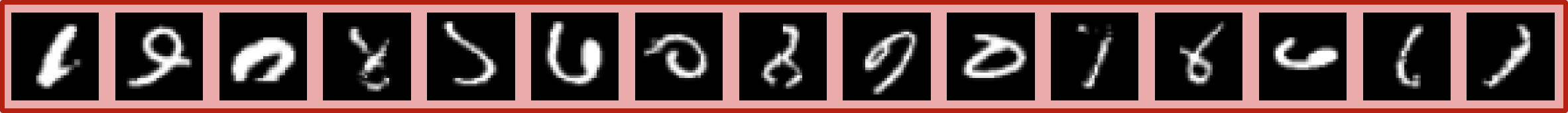}
    \vspace{-0.15cm}
    \caption{High epistemic explanatory uncertainty.}
    \vspace{0.1cm}
    \end{subfigure}
    
    \begin{subfigure}{\textwidth}
    \centering
    \includegraphics[scale=0.28]{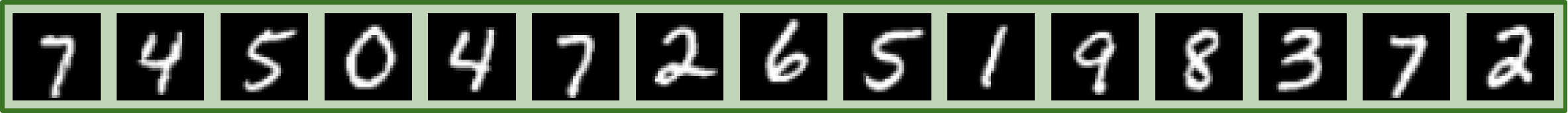}
    \vspace{-0.15cm}
    \caption{Low epistemic explanatory uncertainty.}
    \end{subfigure}
    \label{fig:mnist_high_expl_unc_imgs}
    \caption{
    Examples of inputs receiving (a) high or (b) low explanatory epistemic uncertainty from a Prob-PSENN trained on the MNIST dataset. 
    }
    \label{fig:examples_unc_images}
\end{figure}

\paragraph{Epistemic uncertainty}
Figure~\ref{fig:mnist_high_epistemic} compares the epistemic uncertainty achieved by a Prob-PSENN, trained for the MNIST dataset, when classifying 1000 test samples of the MNIST, F-MNIST and K-MNIST datasets. 
As can be seen, when inputs from the F-MNIST and K-MNIST dataset are presented to the model, a very high explanatory epistemic is captured for almost all the inputs, while for the MNIST test data the uncertainty is uniformly distributed. 
Furthermore, Figure~\ref{fig:examples_unc_images} shows examples of MNIST inputs that achieve the lowest and highest uncertainties. As can be seen, even for the MNIST dataset, highly unusual inputs that could be considered out-of-distribution instances can be found, which are effectively detected by the uncertainty estimation capabilities of Prob-PSENN.

\begin{figure}[!b]
    \centering
    \includegraphics[scale=0.57]{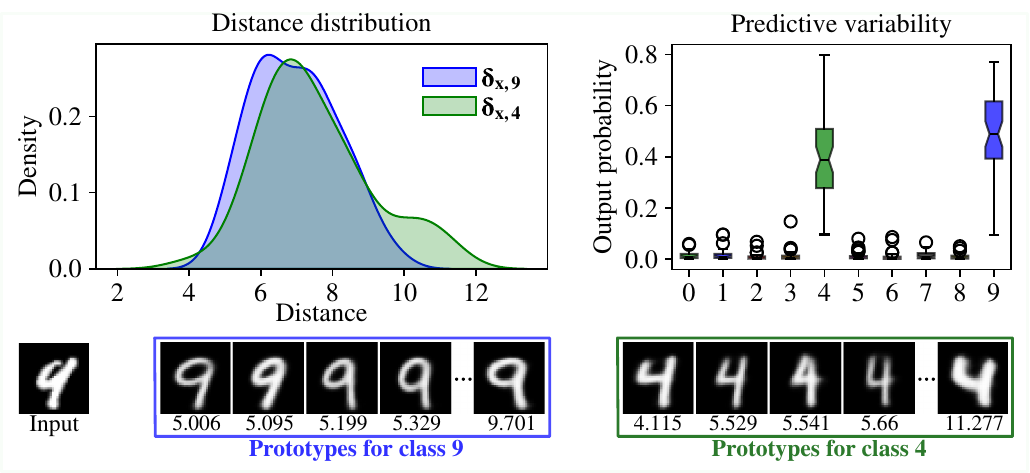}
        \label{fig:expl_protos}
    \caption{
    Prediction and explanation results obtained for a test input with high aleatoric uncertainty. The prototypes are sorted by distance to the input, in increasing order.
    }
    \label{fig:mnist_high_aleatoric}
\end{figure}

\paragraph{Aleatoric uncertainty}
Finally, the prediction and explanation for a particular input with high aleatoric uncertainty is shown in Figure \ref{fig:mnist_high_aleatoric}. The top row includes the density function over the distances corresponding to the two most likely classes, $\*\delta_{x,4}$ and $\*\delta_{x,9}$, estimated by Gaussian-kernel density estimation, as well as a box plot showing the variability in the output probabilities assigned to each class, considering all the inferences. 
Note that the distributions of both the distances and the output probabilities are highly overlapping, evidencing a high explanatory and predictive aleatoric uncertainty. The second row of the figure includes the input and a subset of the sampled prototypes, sorted by distance to the input in increasing order. As can be seen, both distributions over the prototypes capture and sample prototypical representations that could be taken as equally descriptive of the input at hand, thus explaining its ambiguity.

\subsection{Comparison between conventional PSENN and Prob-PSENN}
\label{sec:performance}
\begin{figure}[b]
    \centering
    \includegraphics[scale=0.33]{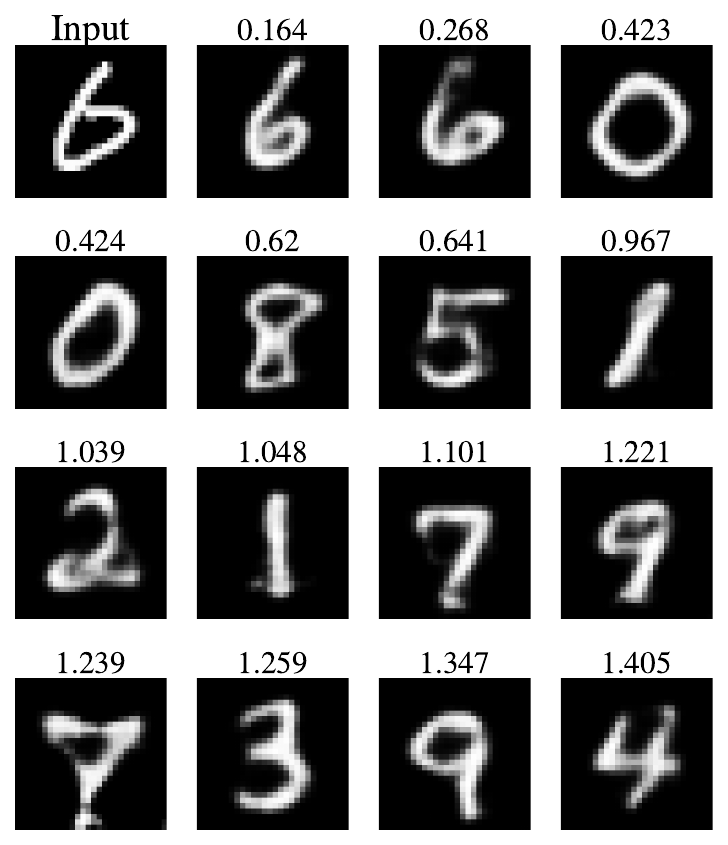}
    \hspace{1.8cm}
    \includegraphics[scale=0.45, trim=0 60 30 10, clip]{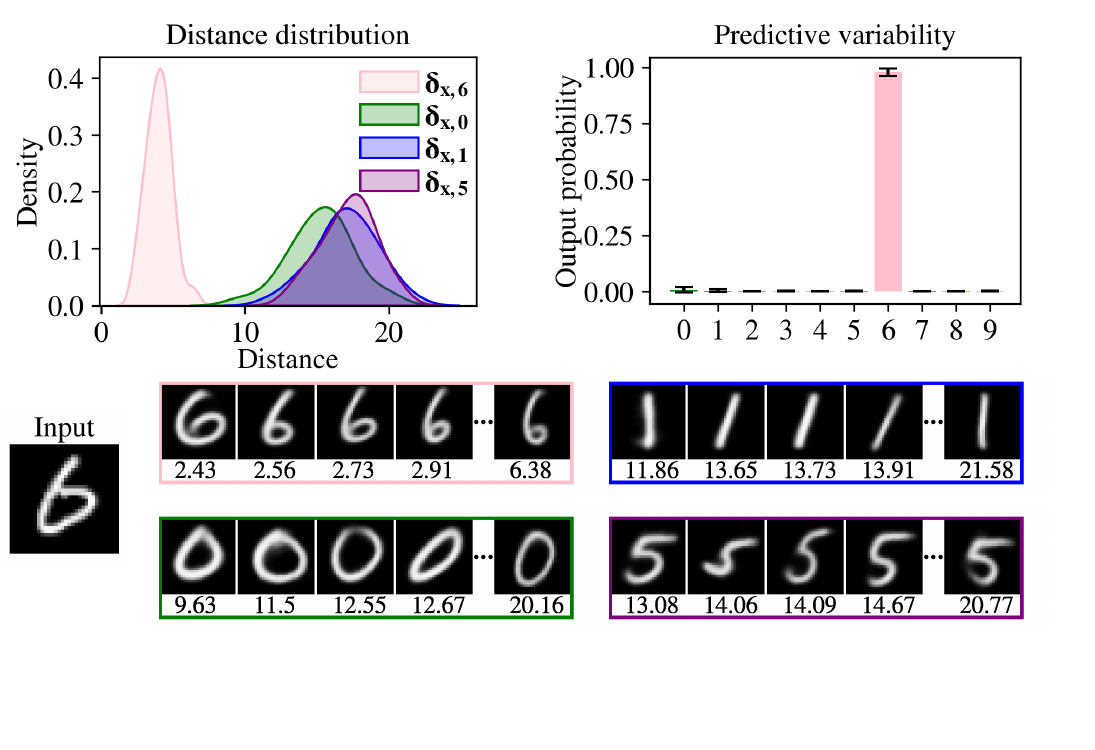}
    \caption{Comparison between the explanation provided by a PSENN \cite{li2018deep} with fixed prototypes (left) and a Prob-PSENN (right). For simplicity, the results corresponding to the four most likely classes are visualized for Prob-PSENN. For each of these classes, we display the four prototypes closest to the input, as well as the most distant prototype.
    }
    \label{fig:comp_deter_prob}
\end{figure}

To analyze the predictive performance of Prob-PSENNs, we begin by comparing the accuracies obtained for the MNIST, F-MNIST and K-MNIST test sets, as shown in Table~\ref{tab:performances}. For reference, we include the results of the deterministic baseline models proposed in Li et al. \cite{li2018deep} and Gautam et al. \cite{gautam2022protovae}, reproduced using the implementation released by the authors (see Appendix \ref{app:related_work} for a further discussion on these works). Results on the SVHN dataset will be reported in Table \ref{tab:acc_svhn} (see Section \ref{sec:scalability}), using a complementary set of baselines more competitive for color images. The results show that Prob-PSENNs achieve either superior or competitive performance in all the datasets, while having as benefits greater trustworthiness and enhanced explanatory capabilities. These capabilities are showcased in Figure~\ref{fig:comp_deter_prob}, which compares, for the MNIST task, the explanation computed by a conventional PSENN ($m=15$) with that computed by Prob-PSENN. More comparisons are provided in Appendices \ref{app:fashion} and \ref{app:kmnist}. As can be seen, the prototypes sampled by Prob-PSENN exhibit considerably more diversity than the fixed prototypes learned by a conventional PSENN, which enables one to find more tailored prototypes for the input at hand, as well as to provide alternative forms that the class might take. In addition, while PSENNs only return point distances to the prototypes, Prob-PSENNs provide a distribution over those distances, enabling assessment of the uncertainty of the prediction, and, hence, accounting for a more trustworthy approach.

\begin{table}[t]
    \centering
    \caption{Predictive accuracies (mean and standard deviation for 5 runs) obtained for the MNIST, Fashion-MNIST and K-MNIST datasets. The accuracies are shown in percentages.
    Best results in the default setting are in bold, while the best results in the uncertainty-aware setting (\textit{Unc. Aware}; see Section \ref{sec:unc_aware_performance}) are in bold and italics.
    }
    \label{tab:performances}
    \scalebox{0.85}{
\begin{tabular}{@{}llccc@{}}
\toprule
Model & Configuration & MNIST & F-MNIST & K-MNIST  \\
\midrule
Li et al. \cite{li2018deep} & ($m=10, l=40$) &  99.0 \std{0.16} & 90.2 \std{0.33} &  91.1 \std{0.47} \\
Li et al. \cite{li2018deep} & ($m=15, l=40$) &  90.0 \std{0.11} & 90.1 \std{0.61} &  90.9 \std{0.41} \\
Li et al. \cite{li2018deep} & ($m=30, l=40$) &  99.1 \std{0.15} & 89.9 \std{0.33} &  91.0 \std{0.96} \\
Li et al. \cite{li2018deep} & ($m=50, l=40$) &  99.0 \std{0.11} & 89.8 \std{0.44} &  91.2 \std{0.74} \\
Gautam et al. \cite{gautam2022protovae} & ($m=50, l=256$) & \textbf{99.4 \std{0.10}} & 91.9 \std{0.20} & 96.1 \std{0.16} \\

\midrule
\textbf{Prob-PSENN} & ($m\!=\!10$, $l\!=\!20$) & \textbf{99.4 \std{0.05}} & 91.9 \std{0.14} & \textbf{96.7 \std{0.21}}  \\

\textbf{Prob-PSENN} & ($m\!=\!10$, $l\!=\!40$) & \textbf{99.4 \std{0.02}} & \textbf{92.0 \std{0.24}} & 96.6 \std{0.17}  \\

\ \ \ + \textit{Unc. Aware} & ($\alpha=0.99$) &  \textitbf{99.9 \std{0.02}}   &  \textit{95.6 \std{0.35}}   &  \textit{99.5 \std{0.03}}  \\ 
\ \ \ + \textit{Unc. Aware} & ($\alpha=0.95$) &  \textitbf{99.9 \std{0.01}}   &  \textitbf{97.3 \std{0.21}}   &  \textitbf{99.7 \std{0.02}}  \\ 

\bottomrule
\end{tabular}
}
\end{table}

\subsubsection{Harnessing uncertainty to improve predictive accuracy and reliability}
\label{sec:unc_aware_performance}
\begin{figure}[]
    \centering
    \includegraphics[scale=0.6]{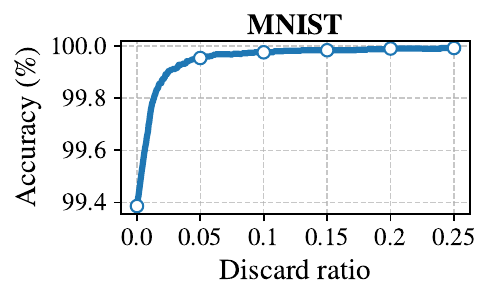}
    \includegraphics[scale=0.6]{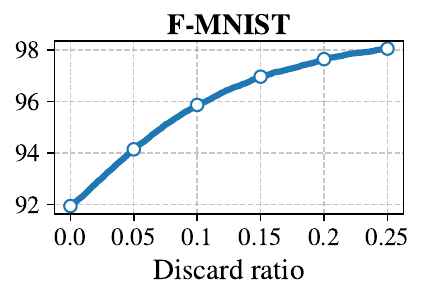}
    \includegraphics[scale=0.6]{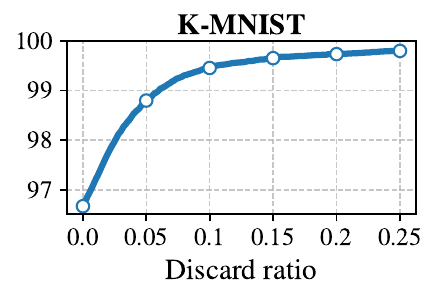}
    \caption{Trade-off between the accuracy gains and the discarded ratio, when the inputs with highest epistemic uncertainty are rejected. Results averaged for 5 runs. Notice that the scales of the Y-axis differ in the three plots.}
    \label{fig:acc_unc_tradeoff}
\end{figure}

We experimentally validated that taking model uncertainty into consideration can provide substantial improvements in accuracy. Specifically, by discarding inputs with high explanatory epistemic uncertainty\footnote{We justify the choice of the mentioned uncertainty metric based on the fact that it is reasonable to consider a prediction unreliable if all the prototypes have low similarity to the input, given the distance-based classification pipeline of the model.} (e.g., greater than $\alpha=95$\% or $\alpha=99$\%), the accuracy of Prob-PSENN increases in all the datasets and configurations, as we show in Table \ref{tab:performances} (last two rows). To complement these results, Table \ref{tab:unc_pred_discarded} shows the percentage of inputs that are discarded in each case. The trade-off between the accuracy and the ratio of discarded samples (i.e., those with highest epistemic uncertainty) is also shown in Figure \ref{fig:acc_unc_tradeoff}.
At the same time, a large number of the discarded inputs are anomalous or challenging to classify, which would justify discarding them for the sake of higher reliability. As evidence, we show in Table \ref{tab:unc_pred_discarded} that the error rate of the deterministic baseline model is dramatically higher in the inputs which receive high uncertainty than in the rest. For instance, the average error rate percentage of the model in the discarded inputs ($\alpha=0.99$) is roughly 23\%, 40\% and 51\% for MNIST, F-MNIST and K-MINST, respectively. 
Thus, this evaluation shows how capturing model uncertainty provides greater accuracy, reliability and safety. 

\begin{table}[b]
    \centering
    \caption{ 
    Percentage of inputs discarded by Prob-PSENNs at different uncertainty thresholds ($\alpha$) and the error rates (\%) of the deterministic baseline model \cite{li2018deep} for high-uncertainty (discarded) and low-uncertainty (retained) inputs across MNIST, F-MNIST, and K-MNIST.
    }
    \label{tab:unc_pred_discarded}
    \scalebox{0.9}{
\begin{tabular}{@{}llccc@{}}
\toprule
Threshold & Metric & MNIST & F-MNIST & K-MNIST \\
\midrule 
$\alpha=0.999$ & 
Discarded Inputs & \ \ 1.38 \std{0.10}   &  \ \ 2.23 \std{0.49}   & \ \ 5.84 \std{0.21}  \\ 
& 
Error Rate (High-Unc.)   &  31.30 \std{3.22}   &  45.49 \std{2.33}   &  64.20 \std{2.61}  \\ 
 & 
Error Rate (Low-Unc.)   &  \ \  0.79 \std{0.20}   &  \ \  9.05 \std{0.57}   &  \ \  6.28 \std{0.98}  \\ 
\midrule 
$\alpha=0.99$ & 
Discarded Inputs & \ \ 3.02 \std{0.03}   &  \ \ 9.19 \std{0.73}   &  11.22 \std{0.23}  \\ 
&
Error Rate (High-Unc.)   &  23.72 \std{2.86}   &  40.01 \std{1.73}   &  50.95 \std{2.73}  \\ 
& 
Error Rate (Low-Unc.)   &  \ \  0.51 \std{0.18}   &  \ \  6.80 \std{0.46}   &  \ \  4.44 \std{0.85}  \\ 

\midrule 
$\alpha=0.95$ & 
Discarded Inputs & \ \ 6.78 \std{0.08}   &  17.52 \std{0.70}   &  19.84 \std{0.28}  \\ 
&
Error Rate (High-Unc.)  &  13.15 \std{1.68}   &  33.28 \std{1.56}   &  36.59 \std{2.75}  \\ 
 & 
Error Rate (Low-Unc.)   &  \ \  0.35 \std{0.16}   &  \ \  4.87 \std{0.34}   &  \ \  3.00 \std{0.63}  \\ 
\midrule 
\end{tabular}
}
\end{table}

\subsubsection{Ablation studies}
\label{sec:ablation}
To isolate the impact of the probabilistic reformulation of the prototypes, Appendix \ref{app:ablation} compares the performance of Prob-PSENN with its deterministic counterpart, using the same architecture and training pipeline. The point-wise prototypes have been optimized using the loss function proposed in \cite{li2018deep}. The results obtained with deterministic PSENNs are summarized in Table \ref{tab:ablation_determ}, computed considering $l\!\in\!\{20,40\}$ latent dimensions and $m\!\in\!\{10,30\}$ prototypes. The results for Prob-PSENN are included in the last rows of the table, as reference. As can be observed, both approaches achieve nearly identical performance, with average accuracy differences below 0.2\%, further demonstrating that the advantages of our probabilistic model do not imply a reduction in its predictive performance.

Furthermore, Table \ref{tab:ablation_determ} in Appendix \ref{app:ablation} also compares Prob‑PSENN with black-box baselines that share the same encoder architecture, followed by a linear classifier trained end-to-end with cross-entropy loss. To ensure a fair comparison, we also evaluated a variant using the full VAE backbone (encoder and decoder), matching the regularization and training dynamics of Prob‑PSENN's autoencoder, even if the decoder plays no functional role in the black-box classification process at inference time. Consistent with prior work \cite{li2018deep,gautam2022protovae}, our results show that Prob‑PSENN matches or outperforms these black-box counterparts. We attribute this to the structured clustering induced by the prototype distributions in the latent space and the robustness gained through averaging predictions over multiple sampled prototypes, reinforcing the practical advantages of Prob‑PSENN's pipeline.

In addition, Appendix \ref{app:ablation} further examines the benefits of using a VAE over a vanilla AE as the backbone architecture. Table \ref{tab:ablation_vae} in the appendix compares the predictive performance of Prob-PSENNs when replacing the VAE with an AE while keeping all other architectural and training details unchanged. The results show that AEs lead to a performance drop ranging from approximately 0.1\% to 1.2\%, depending on the dataset. In addition, Figure \ref{fig:ablation_prototypes} (Appendix \ref{app:ablation}) visualizes randomly sampled prototypes from Prob-PSENNs with different backbone architectures, showing that while vanilla AE backbones provide comparable results in lower latent dimensions (e.g., $l=10$), they struggle to maintain high-quality prototype distributions as the dimensionality increases. In contrast, VAEs enable effective scaling to higher-dimensional spaces. This demonstrates that, while Prob-PSENNs are not inherently restricted to VAE backbones, more regularized latent spaces enhance both predictive performance and prototype quality in high-dimensional settings.

\subsection{Scalability}
\label{sec:scalability}
\input{tables/svhn_acc}
In order to show that our approach can scale to larger numbers of classes, Appendix~\ref{app:emnist} includes results on the E-MNIST dataset \cite{cohen2017emnist}, an extension of MNIST 
which contains handwritten representations of numbers and letters and is composed of 47 classes. 
As can be seen, Prob-PSENN achieves highly diverse and representative sets of prototypes for all the classes, which would require an excessively large number of pointwise prototypes to achieve using deterministic approaches.
Furthermore, to show that Prob-PSENNs can scale to colour datasets, we tested our model on the SVHN dataset. The accuracy of the model in this dataset is reported in Table \ref{tab:acc_svhn}. The results show that the predictive performance of Prob-PSENNs is on par with previous deterministic approaches, when not superior.

\subsection{Application to tabular data}
\label{sec:tabular}
Although prototype-based self-explainable neural networks are often evaluated only on image data \cite{li2018deep, gautam2022protovae, chen2019this, hase2019interpretable, joo2023distributional, wang2023mixture}, in this section, we show that Prob-PSENN can be applied naturally to tabular settings. We demonstrate this using the \textit{Sensorless Drive Diagnosis (SDD)} dataset \cite{bayer2013sensorless}, which contains 58,509 examples of electric motor signals represented by 48 real-valued features and classified into 11 operating conditions. 
We trained Prob-PSENN using the same pipeline described for images, leveraging a three-layer MLP as an encoder with batch normalization and LeakyReLU activations. The decoder mirrors this structure in reverse. 

\begin{figure*}[t]
    \centering
    \includegraphics[scale=0.415]{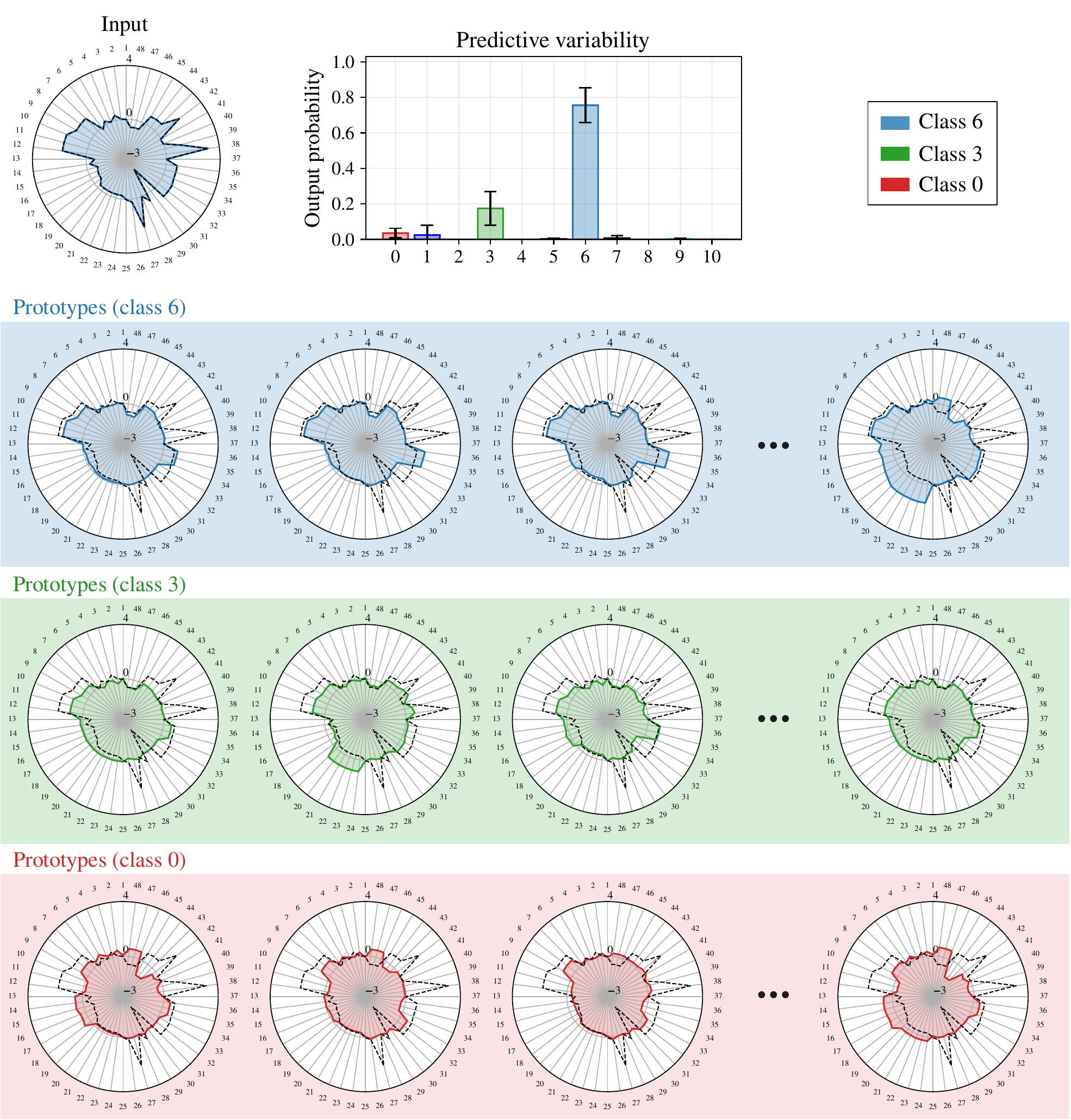}
    \caption{Explanation of a prediction made by Prob-PSENN on an input from the \textit{Sensorless Drive Diagnosis} tabular dataset. The first row contains the input (left) and the class probabilities predicted by Prob-PSENN (middle). 
    The next three rows represent class prototypes sampled from Prob-PSENN, considering, for simplicity, the three most likely classes. Each of these rows displays the three closest prototypes (left) and the most distant one (right), visualized as colored radar plots. The outline of the input (black, dashed line) is overlaid on each prototype radar plot, for direct comparison.
    Overlaps between the input and class prototypes highlight shared patterns, while mismatches reveal deviations or missing features, offering insights into the model’s reasoning, uncertainty, and potential counterfactuals.}
    \label{fig:tabular_explanation}
\end{figure*}

Figure~\ref{fig:tabular_explanation} shows an example explanation produced by a Prob-PSENN trained on the SDD dataset. 
Given the non-visual nature of the data, we leverage a domain-appropriate visualization strategy based on radar plots to visualize the input and the prototypes.
The top row includes the radar plot corresponding to the input sample, alongside the predictive probability distribution over the classes. The three remaining rows include prototypes corresponding to the three most likely classes, in descending order of confidence. Each of these rows contains the three closest prototypes on the left and the farthest prototype on the right. Notice that each prototype is visualized as a color radar chart (solid lines), with the shape of the input overlaid for comparison (dashed lines). The overlapping areas reveal which sensor patterns in the original input match the class fingerprints (large overlap) and which ones deviate (visible gaps).

As can be seen, the explanation highlights both prototypical and divergent feature patterns, helping to understand the input classification, which patterns are shared with other classes, and which features drive uncertainty or ambiguity. These visual explanations show how Prob-PSENN can capture class-representative structures even in tabular domains, broadening the applicability of our model.

\section{Conclusions}
In this paper we introduced Prob-PSENN, a novel Prototype-Based Self-Explainable Neural Network paradigm relying on a probabilistic generalization of the prototypes, which enhances the capabilities of the model in two key aspects. First, unlike previous approaches, our model is not restricted to a fixed number or set of prototypes. This enables Prob-PSENN to learn much more diverse prototypes, and to consider all of them when classifying an input, which represents a more flexible and robust approach. Second, our probabilistic approach equips Prob-PSENNs with tools to capture the uncertainty in both the prediction and the explanation of the network. This novel feature makes it possible to detect when the model is providing uncertain outcomes, substantially increasing the reliability of the model.
We experimentally demonstrated how Prob-PSENNs provide more informative and trustworthy explanations than their non-probabilistic counterparts, while achieving superior or competitive predictive accuracies. Moreover, we show that the accuracy of Prob-PSENNs can be improved by leveraging their novel uncertainty-quantification capabilities, further enhancing their reliability.
Finally, we demonstrate that Prob-PSENN extends naturally to tabular data, enabling interpretable decision-making beyond vision tasks.

\section{Limitations and future work}
As future work, inspired by recent advances in Bayesian Neural Networks \cite{neal2012bayesian,gal2016uncertainty,goan2020bayesian}, we plan to develop a Bayesian formulation for the distribution over the prototypes, which might enable a more theoretically-grounded approach for uncertainty quantification. Furthermore, we also believe that a fully-probabilistic model (i.e., placing a probability distribution on the rest of the network parameters as well) could enhance the quantification of the uncertainty in Prob-PSENNs, or even induce new notions of uncertainty, such as the \textit{encoding} or \textit{class-representation uncertainty}.
Apart from that, while Prob-PSENNs have demonstrated high predictive and explanatory performance, we found it challenging to model probability distributions over the prototypes in very high-dimensional spaces, and, therefore, resort to relatively low dimensional latent spaces in our experiments. We plan to address this limitation in the future, in order to scale the proposed approach to higher-dimensional datasets. Finally, in future work, we plan to extend Prob-PSENN to more complex data modalities, including discrete or heterogeneous tabular datasets that combine categorical and numerical features, as well as explore its applicability to other structured and unstructured data domains.

\section*{Acknowledgements}
MK received partial support from ELSA: European Lighthouse on Secure and Safe AI project (Grant No. 101070617 under UK guarantee) and the ERC under the European Union’s Horizon 2020 research and innovation program (FUN2MODEL, Grant No. 834115). JV, RS and JAL acknowledge support from the Spanish Ministry of Science, Innovation and Universities (Project PID2022-137442NB-I00 and BCAM Severo Ochoa Accreditation CEX2021-001142-S), and the Basque Government (Grant Nos. KK-2024/00030 and IT1504-22).

\bibliographystyle{abbrvnat}
\bibliography{references}

\appendix

\setcounter{table}{0}
\renewcommand{\thetable}{\thesection.\arabic{table}}
\renewcommand{\thefigure}{\thesection.\arabic{figure}}

\newpage

\section{Discussion on previous PSENN approaches}
\label{app:related_work} 
As introduced in Section~\ref{sec:intro}, the development of PSENNs arises as an alternative to black-box DNNs and posthoc explanations, with the goal of ensuring a human-understandable decision process by construction, while maintaining competitive predictive performance. This is achieved by integrating a transparent case-based reasoning as the core element, so that the prediction of the model is determined based on the similarity between the input and a set of prototypical representations of the output classes (see Section~\ref{sec:senn_intro} for further details). 

The way in which such prototypes are defined has led to different types of PSENNs in the literature. 
In the seminal work of Li et al. \cite{li2018deep}, mirrored by recent follow-up works \cite{gautam2022protovae,joo2023distributional}, the prototypes represent complete characterizations of each class or concept (e.g., full representations of the digits in handwritten digit recognition).
In contrast, other follow-up works aim to capture more abstract \cite{alvarez-melis2018robust} or specific features (e.g., prototypical parts or patches) \cite{chen2019this, hase2019interpretable,nauta2021neural} as prototypes.
However, an important drawback of these approaches is that the prototypes are restricted to being samples or sample parts of (encoded) training inputs \cite{alvarez-melis2018robust,chen2019this, hase2019interpretable,nauta2021neural}.
This implies a flexibility loss regarding the end-to-end optimization of the prototypes, as well as a transparency loss when \textit{surrogate prototypes} (e.g., nearest training images) are selected to visualize the explanations \cite{alvarez-melis2018robust, gautam2022protovae}.  For these reasons, in order to guarantee a fully transparent and end-to-end trainable architecture, we followed the same spirit as in \cite{li2018deep,gautam2022protovae}, and used the corresponding approaches as references to build our probabilistic framework.

Closer to our approach, concurrent works \cite{joo2023distributional,wang2023mixture} leverage the use of distributions over the prototypes to improve the consistency between image similarity and latent space location \cite{joo2023distributional}
and to increase the representation power of the model \cite{wang2023mixture}.
Contrary to our work, both approaches employ only the means of the learned distributions as prototypes, whereas our sampling-based inference process allows us to sample multiple sets of prototypes, avoiding the constraint of committing to a fixed set, and harnessing all of them for both the explanation and classification.

Furthermore, the approach in \cite{wang2023mixture} follows the strategy of \cite{alvarez-melis2018robust,chen2019this, hase2019interpretable} and relies on surrogate prototypes, replacing the distribution means with nearest training instances, with the corresponding flexibility and transparency loss discussed above.  
Another key difference between our approach and that in \cite{joo2023distributional} is that, in their work, the similarities between the input's distributional embedding and the prototype distributions, which fed the final classification layer, are measured by considering the entire distributions, using distributional similarity metrics. However, as stated in \cite{joo2023distributional}, such similarities neither form nor rely on observable prototypes, which is contrary to the purpose of explaining and justifying the classification. 
Instead, we advocate for a more transparent approach, in which the model predictions are solely derived from the distances to the sampled sets of prototypes, ensuring a complete and transparent justification of model outputs.

\section{Implementation details}
\label{app:implementation}

\begin{figure}[h]
    \centering
    \begin{subfigure}{\textwidth}
    \centering
    \includegraphics[scale=0.79]{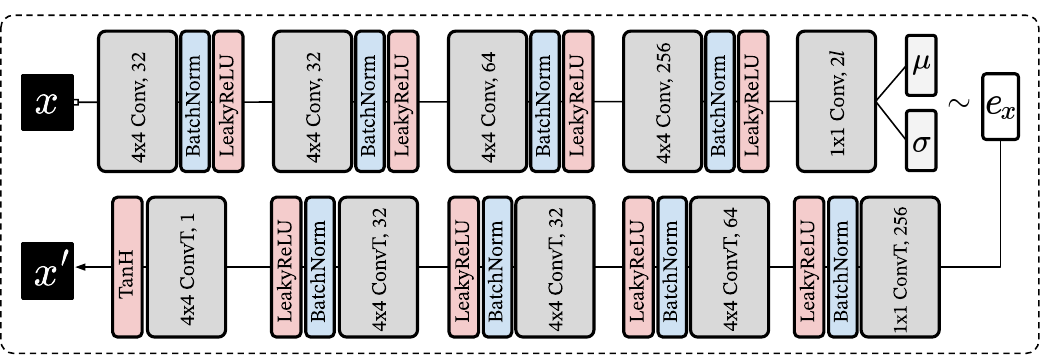}
    \caption{MNIST, F-MNIST, K-MNIST}
    \end{subfigure}
    \begin{subfigure}{\textwidth}
    \centering
    \includegraphics[scale=0.79]{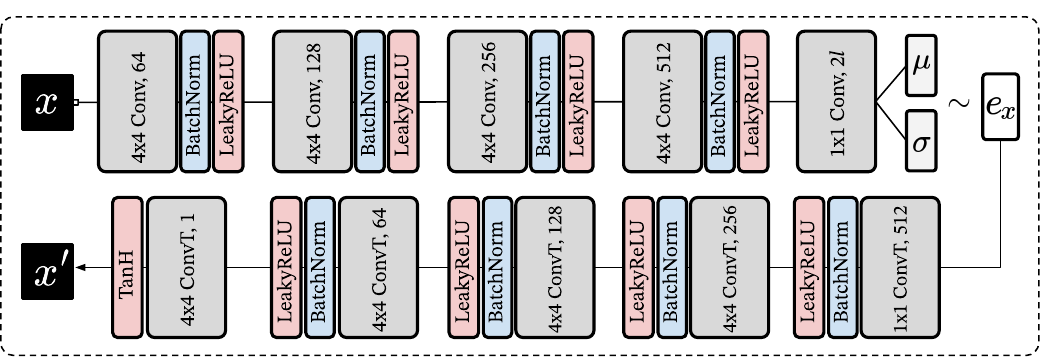}
    \caption{SVHN}
    \end{subfigure}
    \begin{subfigure}{\textwidth}
    \centering
    \includegraphics[scale=0.79]{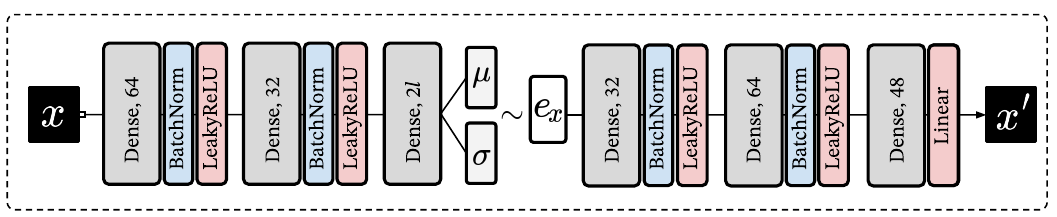}
    \caption{Sensorless Drive Diagnosis}
    \end{subfigure}
    \caption{VAE architectures used in Prob-PSENNs for (a) MNIST, F-MNIST, K-MNIST, (b) SVHN and (c) Sensorless Drive Diagnosis.}
    \label{fig:hyperparameters}
\end{figure}

For image datasets, the Prob-PSENN autoencoder module has been implemented as a CNN-based Variational Autoencoder, consisting of four convolutional layers with $4\times 4$ kernels, batch normalization, and LeakyReLU activations, followed by a $1\times 1$ convolutional layer that controls the dimensionality of the latent space. The decoder mirrors this architecture by means of deconvolutional layers. For tabular data, the architectures consist of three-layer MLPs with batch normalization and LeakyReLU activations. Further details on the architecture and the corresponding hyperparameters are provided in Figure \ref{fig:hyperparameters}. The pixel values have been normalized to the range $[-1,1]$, and, for grayscale datasets, the input shapes have been upscaled from $28 \times 28$ to $32 \times 32$ using bilinear interpolation. For the SVHN dataset, random brightness adjustments and zoom transformations have been incorporated during training to improve generalization. The experiments have been run for $\tau_1=1.0, \tau_2=10.0, \tau_3=0.05, \tau_4=0.05$ and a batch size of 128. The models have been trained for 100 epochs on grayscale datasets and 30 epochs on the SVHN dataset.
The Adam optimization method \cite{kingma2015adam} has been used to train the networks, using a learning rate of $0.001$. The final classification layer $W$ has been set as $W=-I$, where $I$ represents the identity matrix of dimension $c$, thus ensuring that the association between the prototype $\*r_i$ and the class $y_i$, $1\leq i \leq c$, is explicit and unequivocal. 
In order to ensure higher stability during the training process, the value of $\mathcal{L}_{INT}$ has been cut to the range $[-1,10^3]$.
For all datasets, standard training and test partitions have been used, with 10\% of the training data reserved for validation.
The results of \cite{li2018deep,gautam2022protovae} have been reproduced using the implementations released by the authors. Our code is publicly available at \url{https://github.com/vadel/Prob-PSENN}.

\section{Complementary results on the MNIST dataset}
\label{app:mnist_results}
\begin{figure}[!h]
    \centering
    \includegraphics[scale=0.65]{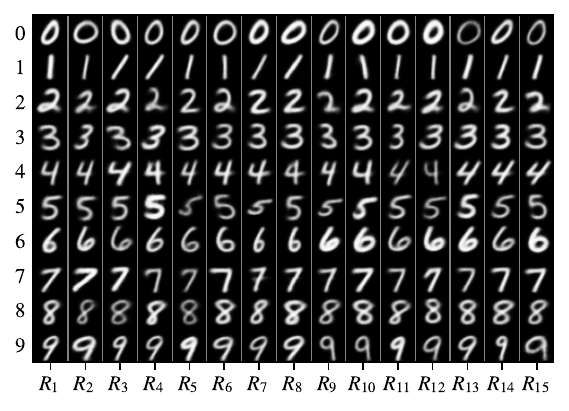}
    \hfill
    \includegraphics[scale=0.65]{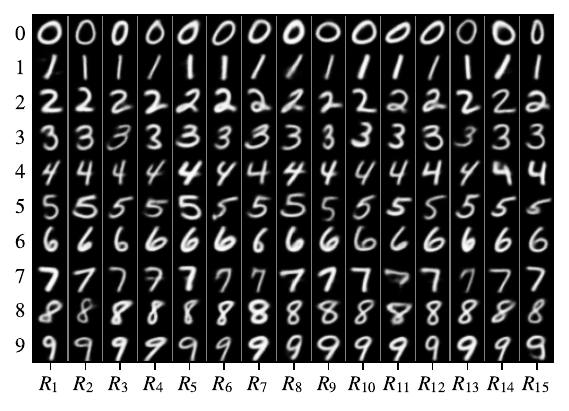}
    \caption{
    15 random sets of prototypes, obtained with a Prob-PSENN with $l=5$ (left) and $l=40$ (right).
    }
    \label{fig:mnist_highed_d_summary}
\end{figure}

In order to complement the results reported in Section \ref{sec:results}, Figure \ref{fig:mnist_highed_d_summary} shows, for the MNIST dataset, 15 randomly sampled sets of prototypes from a Prob-PSENN with $l=5$ (left) and $l=40$ (right). 
As can be observed, the increase in the dimensionality of the latent space makes it possible to capture prototype distributions with a higher degree of complexity in their structure, as well as to obtain more detailed decodings.

\newpage

\section{Results on the Fashion-MNIST dataset}
\label{app:fashion}
\begin{figure}[!h]
    \centering
    \includegraphics[scale=0.6]{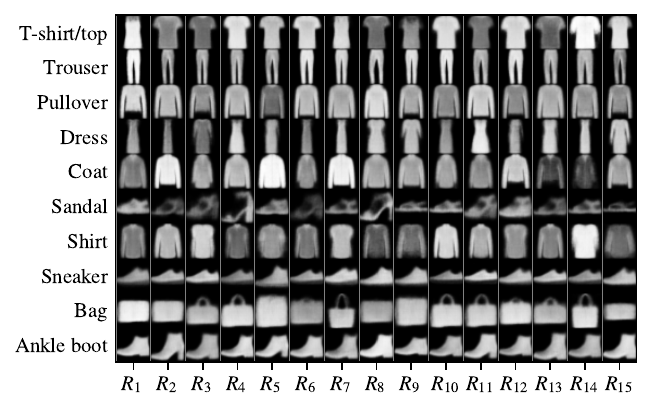}
    \hfill
    \includegraphics[scale=0.6]{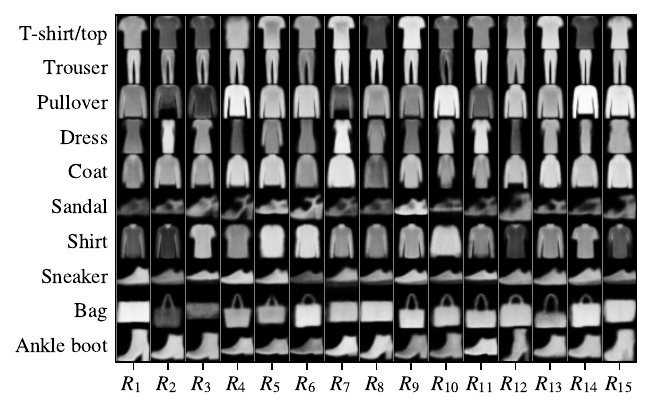}
    \caption{Sampled prototypes for the Fashion-MNIST dataset, using a Prob-PSENN with $l=5$ (left) and $l=40$ (right).
    \\
    }
    \label{fig:fashion_proto_samples}
\end{figure}

This section includes complementary results on the Fashion-MNIST dataset. First, Figure~\ref{fig:fashion_proto_samples} shows randomly sampled prototypes for a Prob-PSENN with $l=5$ (left) and $l=40$ (right). These figures demonstrate that the probabilistic redefinition of the prototypes introduced in Prob-PSENNs enables the model to learn, even for relatively low-dimensional latent spaces, diverse and prototypical representations of each class.\\

\begin{figure}[!h]
    \centering
    \includegraphics[scale=0.485]{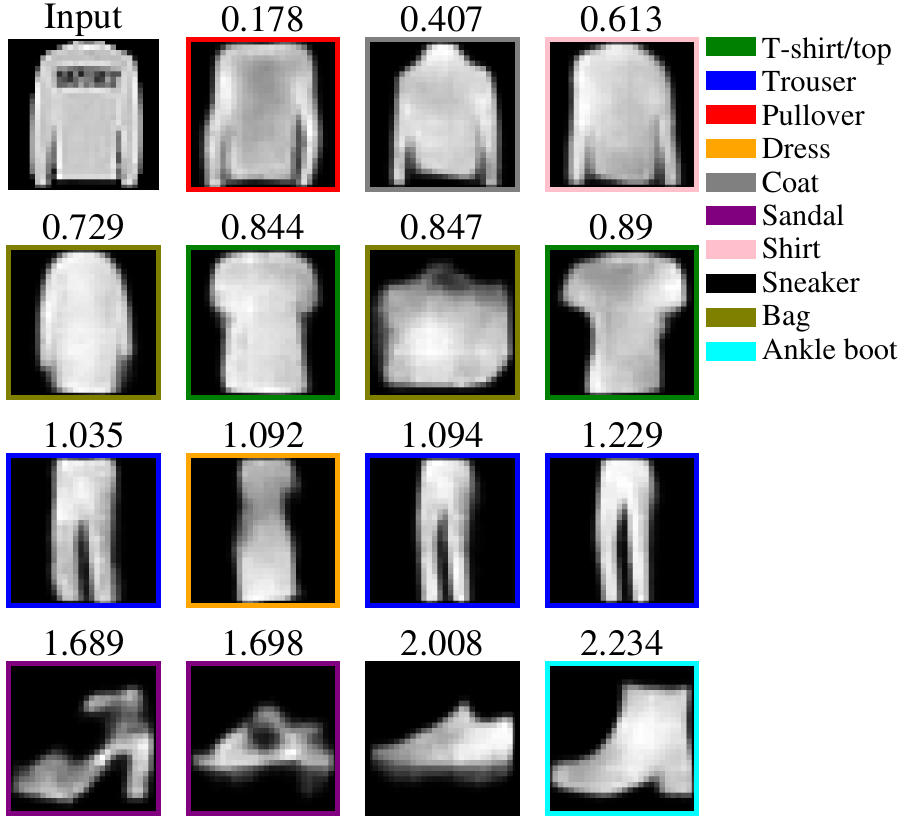}
    \includegraphics[scale=0.47]{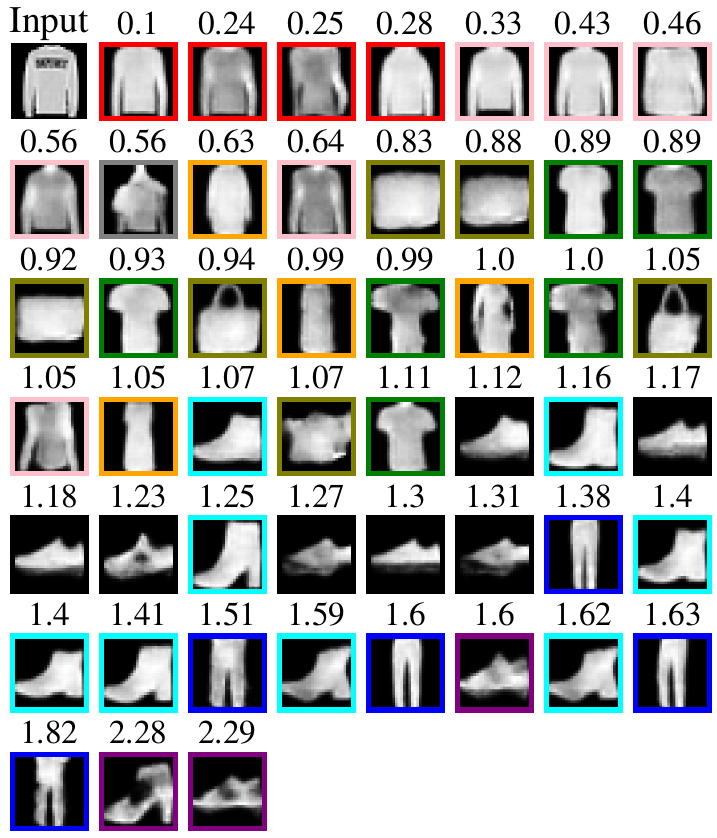}
    \caption{Explanations provided by a PSENN \cite{li2018deep} with $m=15$ (left) and $m=50$ (right) for the Fashion-MNIST dataset.
    Both models classify the input (shown in the top-left corner of each grid) as the class \textit{pullover} with a probability of $0.99$.  
    Note that the prototypes are sorted  based on their distances to the input (displayed on top of each figure), in increasing order.
    \\
    }
    \label{fig:fashion_senn_exp}
\end{figure}

\begin{figure}[!h]
    \centering
    \includegraphics[scale=0.79, trim=0 60 30 0, clip]{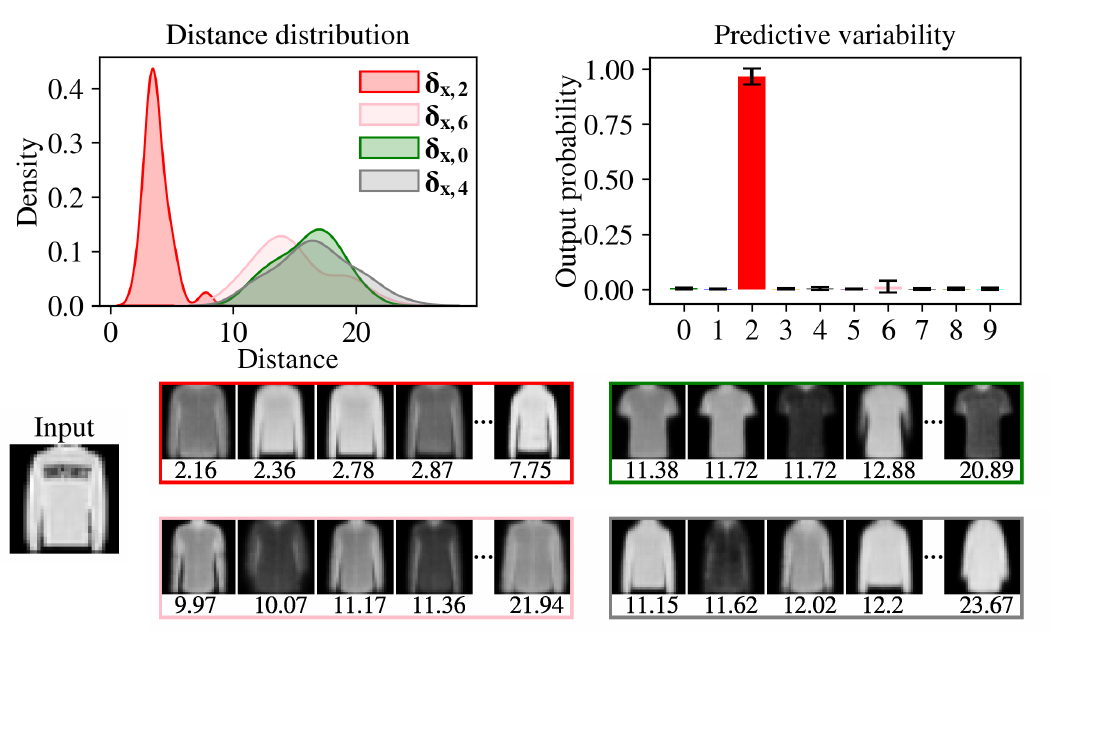}
    \caption{Explanation provided by a Prob-PSENN, for the Fashion-MNIST dataset. The top row shows the distance distribution corresponding to the four most likely classes (left), and the variability in the output probabilities (right). The bottom part displays the prototypes for the four most likely classes, along with their corresponding distances to the input. Note that the prototypes are ordered by their distance, in ascending order.
    \\
    }
    \label{fig:fashion_probsenn_exp}
\end{figure}

Furthermore, a visual comparison between the explanations provided by a non-probabilistic PSENN and a Prob-PSENN is provided in Figures~\ref{fig:fashion_senn_exp} and \ref{fig:fashion_probsenn_exp}. For the sake of completeness, the former figure includes the explanations provided by a PSENN with $m=15$ (left) and $m=50$ (right) prototypes. In both cases, the original input is shown in the top-left corner, and the prototypes are colored according to the classes returned by the model when the prototypes are classified, in order to make it clearer which class each prototype is most associated with. Notice that this connection is explicit and unequivocal by construction in our Prob-PSENNs, being the class $y_i$ represented by the prototype $r_i$, $1\leq i \leq c$, resulting in more transparent decision processes and easier-to-interpret explanations.
As can be observed in Figures~\ref{fig:fashion_senn_exp} and \ref{fig:fashion_probsenn_exp}, the prototypes learned by the PSENNs are, even for $m=50$ prototypes, considerably less meaningful and varied than the ones captured by Prob-PSENN. Furthermore, the uncertainty metrics introduced in Section \ref{sec:explanation_uncertainty} enable us to quantify a low aleatoric and epistemic explanatory uncertainty for this input ($0.004$ and $0.16$, respectively), which increases the trustworthiness and reliability of the outcomes of Prob-PSENN.

\newpage
\vfill

\section{Results on the K-MNIST dataset}
\label{app:kmnist}

\begin{figure}[!h]
    \centering
    \includegraphics[scale=0.44]{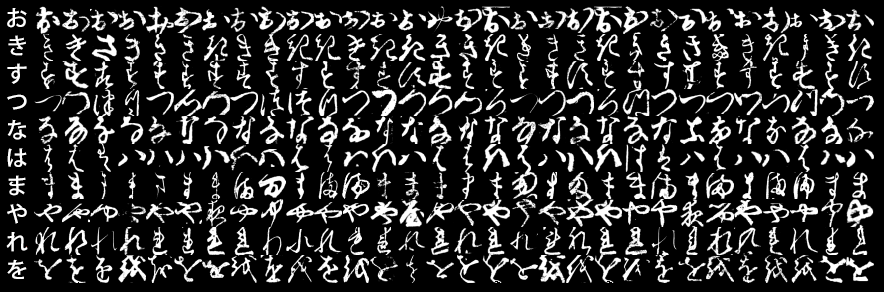}
    \caption{Examples of the K-MNIST dataset \cite{clanuwat2018deep}, as reported in the original repository \url{https://github.com/rois-codh/kmnist} (CC BY-SA 4.0 license). Each row corresponds to one class. The first column of the figure shows the modern representation of each letter, as a reference.}
    \label{fig:kmnist_dataset}
\end{figure}

\begin{figure}[!h]
    \centering
    \includegraphics[scale=0.67]{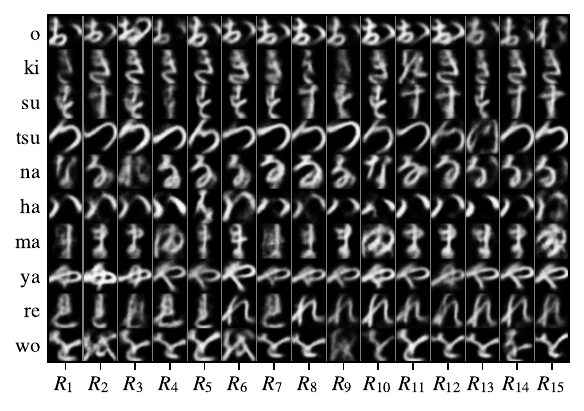}
    \hfill
    \includegraphics[scale=0.67]{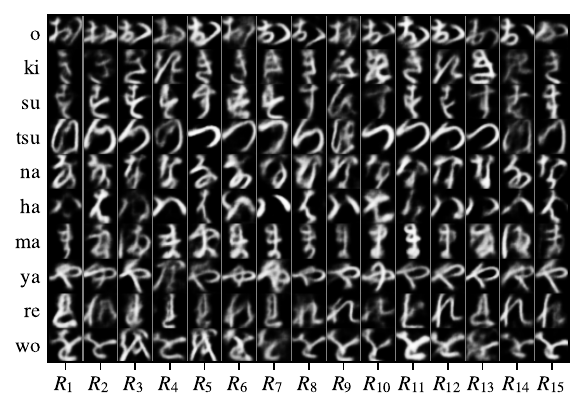}
    \caption{Sampled prototypes for the K-MNIST dataset, using a Prob-PSENN with $l=5$ (left) and $l=40$ (right).}
    \label{fig:kmnist_proto_sample}
\end{figure}

This section contains additional results on the 10-class K-MNIST dataset \cite{clanuwat2018deep}, which includes handwritten representations of Japanese letters. Random images from this dataset are shown in Figure~\ref{fig:kmnist_dataset}, from which it can be noticed a high variance in the representation of each letter. Randomly sampled prototypes from Prob-PSENNs with $l=5$ and $l=40$ are displayed in Figure~\ref{fig:kmnist_proto_sample}. As can be assessed in the figure, despite the complexity and high variability of this dataset, Prob-PSENNs are capable of capturing the most frequent representations of each class, ensuring a sufficiently diverse and meaningful characterization of the classes.

\begin{figure}[!h]
    \centering
    \includegraphics[scale=0.502]{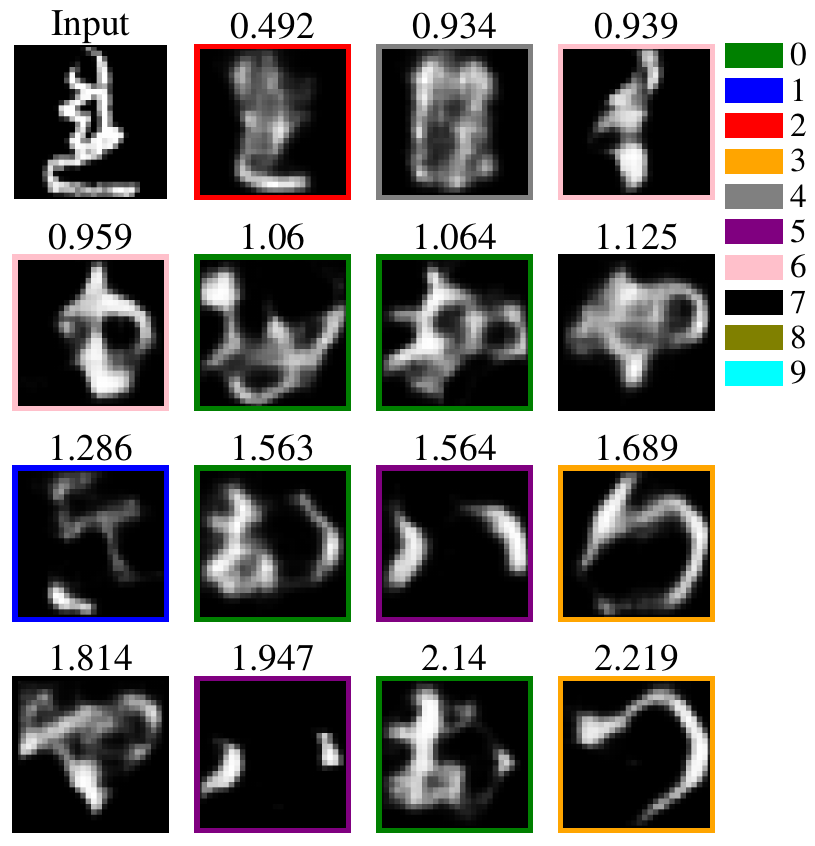}
    \includegraphics[scale=0.5]{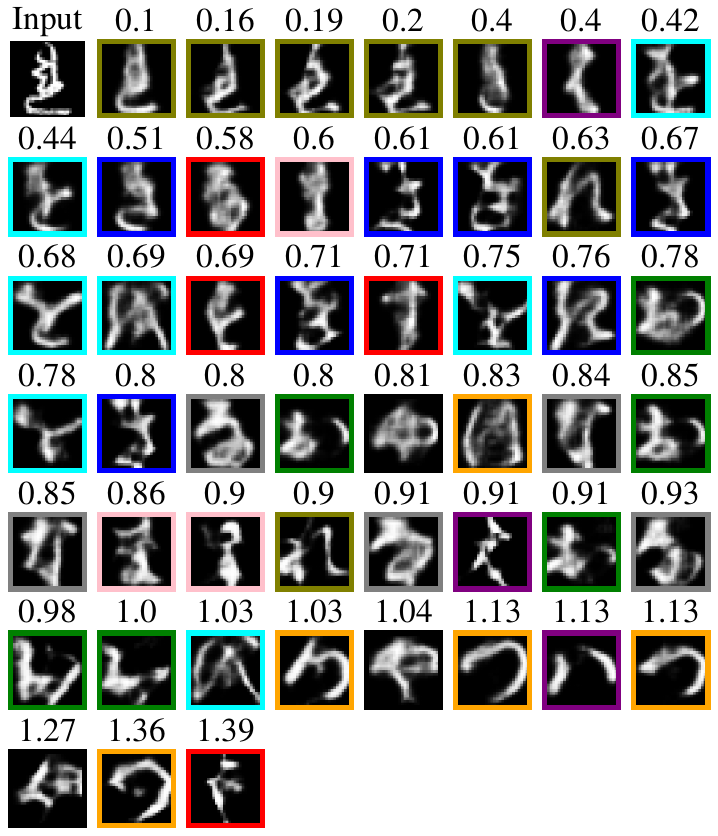}
    \caption{Explanations provided by a PSENN \cite{li2018deep} with $m=15$ (left) and $m=50$ (right) for the K-MNIST dataset. 
    Both models classify the input (shown in the top-left corner of each grid) as the class \textit{8} with a probability of $0.99$.  
    Note that the prototypes are sorted based on their distances to the input (displayed on top of each figure), in increasing order.\\}
    \label{fig:kmnist_senn_exp}
\end{figure}

\begin{figure}[!h]
    \centering
    \includegraphics[scale=0.79, trim=0 60 30 0, clip]{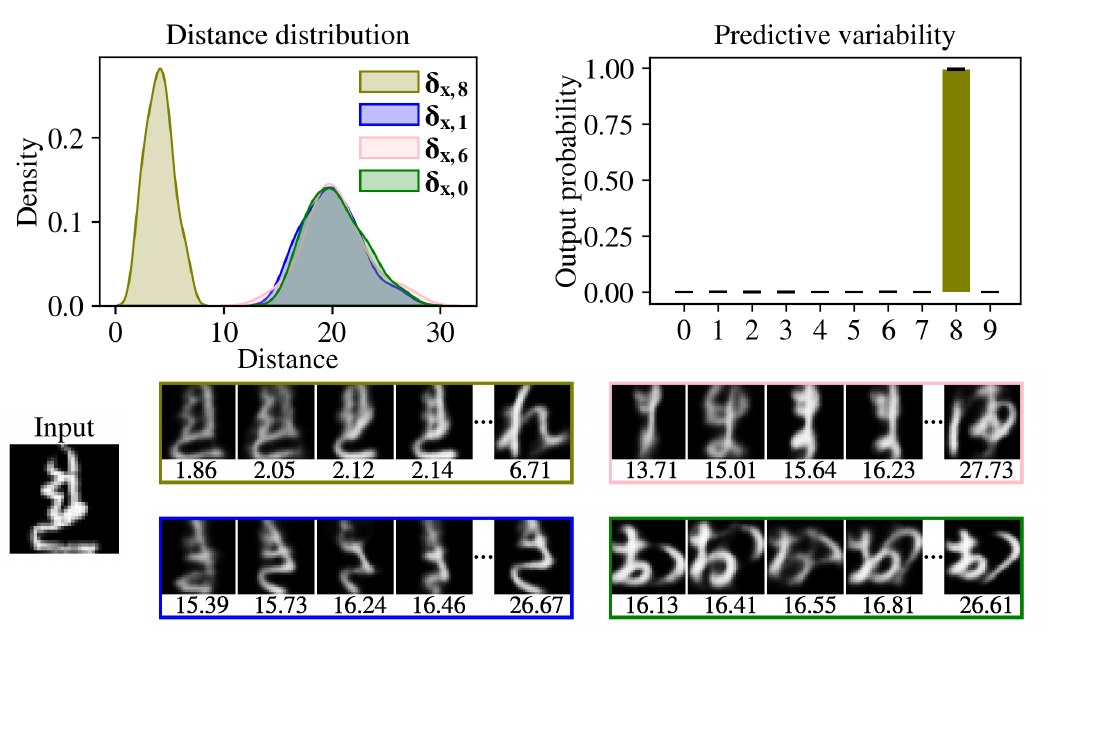}
    \caption{Explanation provided by a Prob-PSENN, for the K-MNIST dataset. The top row shows the distance distribution corresponding to the four most likely classes (left), and the variability in the output probabilities (right). The bottom part displays the prototypes for the four most likely classes, along with their corresponding distances to the input. Note that the prototypes are ordered by their distance, in ascending order.\\}
    \label{fig:kmnist_probsenn_exp}
\end{figure}

As for the previous datasets, a comparison between the explanations of PSENNs and Prob-PSENN for the K-MNIST dataset is provided in Figures~\ref{fig:kmnist_senn_exp} and \ref{fig:kmnist_probsenn_exp}.
Notice how, also due to the higher variability in the representations of each class, the prototypes learned by a PSENN with $m=15$ prototypes (Figure~\ref{fig:kmnist_senn_exp}-left) are not sufficiently representative of the classes, and, consequently, we need to resort to a large number of prototypes ($m=50$, Figure~\ref{fig:kmnist_senn_exp}-right). However, even for $m=50$, the prototypes learned by PSENN are considerably less varied and informative than the ones sampled by Prob-PSENN, which evidences the higher flexibility of our approach, as well as its increased transparency and explanatory capabilities.

\vspace{2cm}

\vfill
\newpage
\pagebreak

\section{Ablation studies: complementary results}
\label{app:ablation}

This section complements the results from Section \ref{sec:performance} by comparing the performance of Prob-PSENN with its (i) black-box and (ii) deterministic counterparts. The black-box models use the same encoder architecture as Prob-PSENN, followed by a linear classifier, and have been trained end-to-end with the cross-entropy loss. For the sake of completeness, and to ensure a fair comparison, we also evaluated a black-box variant using the full VAE backbone (encoder and decoder), matching the regularization and training dynamics of Prob‑PSENN's backbone. Nonetheless, notice that the decoder plays no functional role in the black-box classification process at inference time. The deterministic PSENNs share the same architecture as Prob-PSENN, but replace the probabilistic prototype distributions with a fixed set of $m$ trainable point-wise prototypes. These models have been trained using the classification and reconstruction loss of Prob-PSENN, as well as the interpretability regularizers proposed in \cite{li2018deep}. Both ablation variants have been evaluated considering $l=20$ and $l=40$ dimensions for the latent space, and various numbers of prototypes $m$ for the deterministic PSENNs.

The results obtained with these ablated models are summarized in Table \ref{tab:ablation_determ}. First, we can notice that Prob‑PSENN matches or outperforms these black-box counterparts, a result consistent with prior related works \cite{li2018deep,gautam2022protovae}. We attribute this to the structured clustering induced by the prototype distributions in the latent space and the robustness gained through averaging predictions over multiple sampled prototypes, reinforcing the practical advantages of Prob‑PSENN's pipeline. Next, comparing Prob-PSENN with their deterministic versions, it can be observed that the performance of both approaches is nearly identical, with differences in average accuracy below 0.2\%. These results further demonstrate that the advantages of our probabilistic model do not imply a reduction in its predictive performance, while bringing as benefits enhanced trustworthiness, uncertainty quantification and explanatory capabilities.

\begin{table}[!h]
    \centering
    \caption{Predictive accuracy percentages (mean and standard deviation for 5 runs) for black-box and deterministic PSENNs using the same architecture and training pipeline as Prob-PSENNs.
    The \textbf{best results for each approach} are highlighted in bold, in order to facilitate comparison and emphasize their similar performance.
    }
    \label{tab:ablation_determ}
    \scalebox{0.93}{
\begin{tabular}{@{}llcccc@{}}
\toprule
Model & Configuration & MNIST & F-MNIST & K-MNIST & SVHN \\
\midrule

Black-box (Enc. only) & 
 ($m=0$, $l\!=\!20$) & 
 99.11 \std{0.19} & 
 90.77 \std{0.15} &  
 95.58 \std{0.57} & 
 90.94 \std{0.30} \\

Black-box (Enc. only) & 
 ($m=0$, $l\!=\!40$) & 
 99.00 \std{0.31} & 
 90.59 \std{0.22} &  
 95.63 \std{0.43} & 
 91.10 \std{0.39} \\

Black-box (Enc. + Dec.) & 
 ($m=0$, $l\!=\!20$) & 
 \textbf{99.20 \std{0.03}} & 
 91.66 \std{0.20} &  
 95.83 \std{0.16} & 
 91.53 \std{0.56} \\

Black-box (Enc. + Dec.) & 
 ($m=0$, $l\!=\!40$) & 
 99.17 \std{0.04} & 
 \textbf{91.82 \std{0.16}} &  
 \textbf{95.93 \std{0.18}} & 
 \textbf{91.74 \std{0.41}} \\

\midrule

PSENN & ($m\!=\!10$, $l\!=\!20$) & 99.39 \std{0.04} & \textbf{92.11 \std{0.10}} &  96.71 \std{0.12} & 92.37 \std{0.28} \\

PSENN & ($m\!=\!10$, $l\!=\!40$) & \textbf{99.42 \std{0.02}} & 91.94 \std{0.25} &  \textbf{96.79 \std{0.08}} & \textbf{92.44 \std{0.32}} \\

PSENN & ($m\!=\!30$, $l\!=\!20$) & 99.33 \std{0.10} & 90.92 \std{0.14} &  96.42 \std{0.18} & 91.05 \std{0.41} \\

PSENN & ($m\!=\!30$, $l\!=\!40$) & 99.29 \std{0.08} & 90.30 \std{0.51} &  96.72 \std{0.11} & 90.52 \std{0.23} \\

\midrule
Prob-PSENN & ($m\!=\!10$, $l\!=\!20$) & \textbf{99.42 \std{0.05}} & 91.91 \std{0.14} & \textbf{96.72 \std{0.21}} & 91.62 \std{0.56} \\

Prob-PSENN & ($m\!=\!10$, $l\!=\!40$) & 99.38 \std{0.02} & \textbf{91.98 \std{0.24}} & 96.61 \std{0.17} & \textbf{92.14 \std{0.24}} \\
\bottomrule
\end{tabular}
}
\end{table}

\begin{figure}[]
    \centering
    
    \begin{subfigure}{\textwidth}
    \centering
    \includegraphics[scale=0.5]{img/proto_small/proto_samples_mnist_40.pdf}
    \includegraphics[scale=0.5]{img/proto_small/proto_samples_fashion_40.pdf}
    \includegraphics[scale=0.5]{img/proto_small/proto_samples_hiragana_40.pdf}
    \includegraphics[scale=0.5]{img/proto_small/proto_samples_svhn_40.pdf}
    \vspace{-0.2cm}
    \caption{Prob-PSENN with a \textbf{VAE} backbone, $l=\mathbf{40}$.}
    \vspace{0.2cm}
    \end{subfigure}

    \begin{subfigure}{\textwidth}
    \centering
    \includegraphics[scale=0.5]{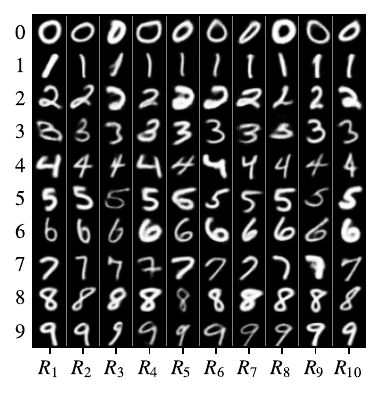}
    \includegraphics[scale=0.5]{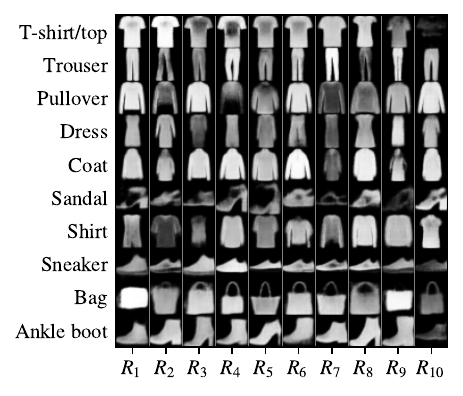}
    \includegraphics[scale=0.5]{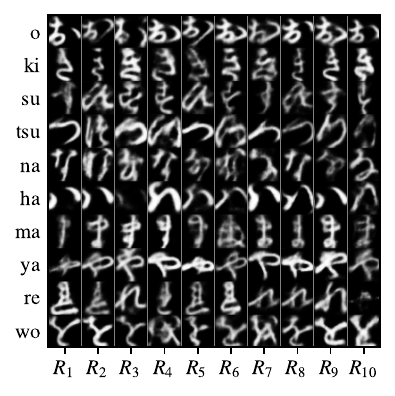}
    \includegraphics[scale=0.5]{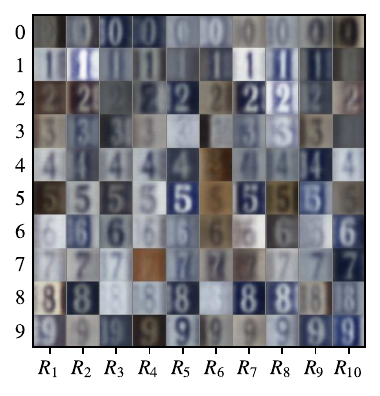}
    \vspace{-0.2cm}
    \caption{Prob-PSENN with an \textbf{AE} backbone, $l=\mathbf{10}$.}
    \vspace{0.2cm}
    \end{subfigure}
    
    \begin{subfigure}{\textwidth}
    \centering
    \includegraphics[scale=0.5]{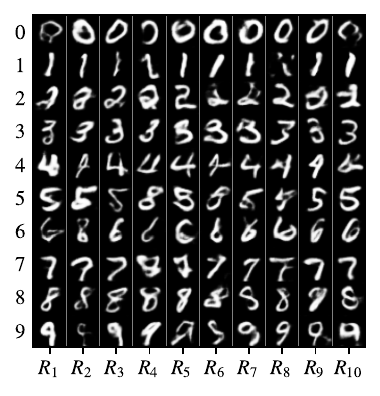}
    \includegraphics[scale=0.5]{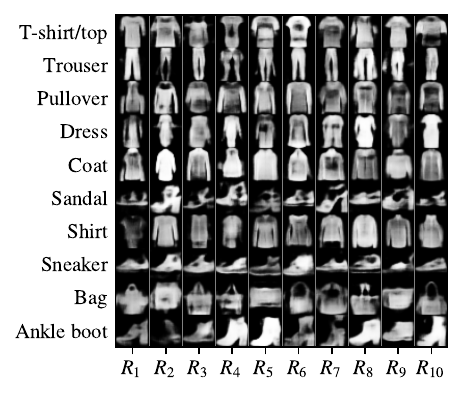}
    \includegraphics[scale=0.5]{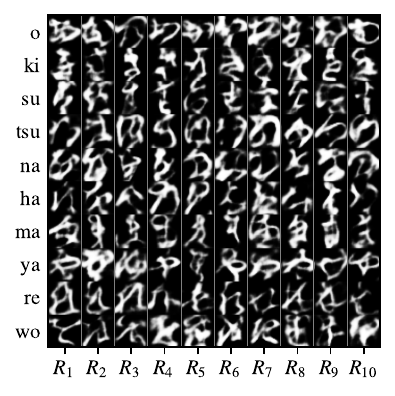}
    \includegraphics[scale=0.5]{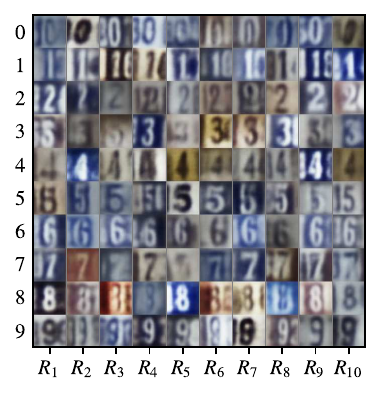}
    \vspace{-0.2cm}
    \caption{Prob-PSENN with an \textbf{AE} backbone, $l=\mathbf{40}$.}
    \end{subfigure}
    
    \caption{
    Visualization of randomly sampled prototype sets from Prob-PSENNs with the following backbone configurations: (a) VAE and $l=40$, (b) AE and $l=10$, and (c) AE and $l=40$. 
    Prob-PSENNs with AE backbones provide a good performance in lower dimensions (e.g., $l=10$), but struggle to learn high-quality prototype distributions as dimensionality increases, whereas VAEs enable effective scaling to higher-dimensional spaces.
    }
    \label{fig:ablation_prototypes}
\end{figure}

Finally, in Table \ref{tab:ablation_vae} we report the results obtained with Prob-PSENNs for which VAEs have been replaced with vanilla AutoEncoders, matching all the remaining architectural details and the training pipeline. The results show that vanilla AutoEncoders lead to a performance drop ranging from approximately 0.1\% to 1.2\%, depending on the dataset. In addition, Figure \ref{fig:ablation_prototypes} visualizes different sets of randomly sampled prototypes from Prob-PSENNs with different backbone architectures, highlighting the advantage of regularized latent spaces for improving prototype quality in high-dimensional settings.

\begin{table}[h]
    \centering
    \caption{Comparison between the accuracy of Prob-PSENNs using vanilla AutoEncoder (AE) and Variational AutoEncoder (VAE) backbones.
    }
    \label{tab:ablation_vae}
\begin{tabular}{@{}llcccc@{}}
\toprule
Model & Configuration & MNIST & F-MNIST & K-MNIST & SVHN  \\
\midrule
Prob-PSENN & (AE,  $l\!=\!10$) & 99.26 \std{0.06}  &  91.21 \std{0.48} & 95.58 \std{0.36}  & 90.91 \std{0.26} \\
Prob-PSENN & (AE,  $l\!=\!20$) & 99.36 \std{0.05} & 91.76 \std{0.21}  & 96.01 \std{0.12} & 90.72 \std{0.41} \\
Prob-PSENN & (AE,  $l\!=\!40$) & 99.35 \std{0.09}  & 91.65 \std{0.30} & 96.00 \std{0.15} & 90.37 \std{0.22} \\

\midrule
Prob-PSENN & (VAE,  $l\!=\!20$) & \textbf{99.42 \std{0.05}} & 91.91 \std{0.14} & \textbf{96.72 \std{0.21}} & 91.62 \std{0.56} \\

Prob-PSENN & (VAE,  $l\!=\!40$) & 99.38 \std{0.02} & \textbf{91.98 \std{0.24}} & 96.61 \std{0.17} & \textbf{92.14 \std{0.24}}  \\
\bottomrule
\end{tabular}
\end{table}

\newpage

\section{Results on the E-MNIST dataset}
\label{app:emnist}

\begin{figure}[t]
    \centering
    \includegraphics[scale=0.6]{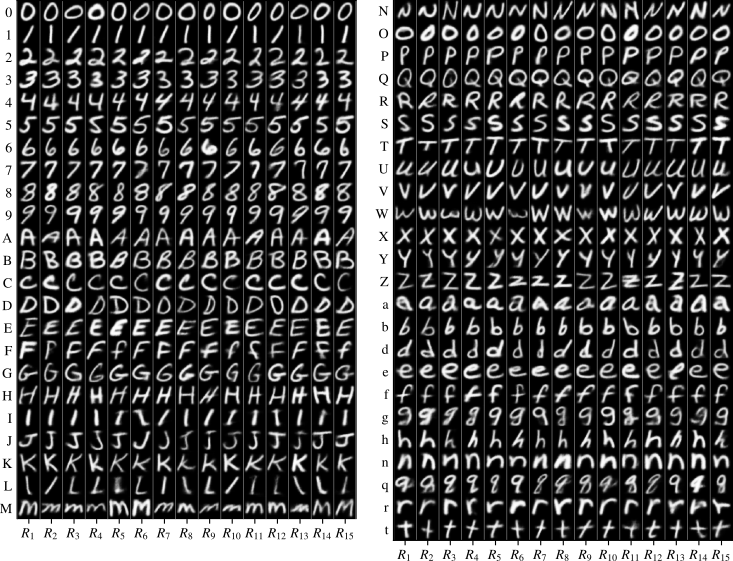}
    \caption{Prototypes sampled for a Prob-PSENN trained on the E-MNIST \cite{cohen2017emnist} dataset (47 classes), consisting of handwritten representations of numbers and letters.}
    \label{fig:emnist_prototypes}
\end{figure}

In order to analyze the scalability of Prob-PSENN in tasks consisting of a larger number of classes, we evaluated its performance in the E-MNIST dataset \cite{cohen2017emnist} (47 classes), an extension of MNIST that includes both numeric and alphabetic characters. As can be seen in Figure \ref{fig:emnist_prototypes}, Prob-PSENN is still capable of learning suitable distributions over the prototypes for the 47 classes, yielding realistic and diverse prototypes, while achieving a test accuracy of 0.89.

\section{Broader Impact}

As specified above, the goal of Prob-PSENNs is to promote a transparent-by-design DNN architecture while maintaining competitive predictive performance. In addition to transparency, Prob-PSENNs provide tools to quantify the uncertainty in the model's predictions and explanations, which further enhances its safety and reliability. Therefore, our work is aligned with three important needs of deep learning models: interpretability, transparency, and reliability. However, a current concern in deep learning models is their lack of robustness to adversarial attacks \cite{yuan2019adversarial,szegedy2014intriguing,biggio2013evasion}, a vulnerability that also threatens explanation methods and self-explainable models \cite{vadillo2025adversarial,aivodji2019fairwashing,lakkaraju2020how,ghorbani2019interpretation}. 
Therefore, in order to ensure a safe and responsible deployment of Prob-PSENNs in practice, it remains crucial to carefully assess the implications of adversarial attacks, and, consequently, to evaluate the robustness of the model against them.

\end{document}

%% file: tables/svhn_acc.tex
\begin{wraptable}{r}{5.5cm}
    \centering
    \caption{Results on the SVHN dataset.}
    \label{tab:acc_svhn}
    \scalebox{0.75}{
\begin{tabular}{@{}lc@{}}
\toprule
Configuration & Accuracy (\%)  \\
\midrule
Gautam et al. \cite{gautam2022protovae} (ProtoPNet) & $88.6 \pm 0.3$ \\
Gautam et al. \cite{gautam2022protovae} (ProtoVAE) & $92.2 \pm 0.3$ \\
Gautam et al. \cite{gautam2022protovae} (SENN) & $91.5 \pm 0.4$ \\
\midrule
\textbf{Prob-PSENN} & $92.1 \pm 0.2 $ \\

\bottomrule
\end{tabular}
}
\end{wraptable}